\documentclass{article}

\usepackage{microtype}
\usepackage{graphicx}
\usepackage{subcaption}
\usepackage{booktabs} % for professional tables
\usepackage{siunitx} % for proper unit formatting
\usepackage{tikz} 
\usetikzlibrary{shapes, arrows.meta, positioning, calc, decorations.pathreplacing}
\usepackage{hyperref}

\usepackage[preprint]{icml2026}

\usepackage{amsmath}
\usepackage{amssymb}
\usepackage{mathtools}
\usepackage{amsthm}

\newcommand{\RETURN}{\STATE \textbf{return} }

\usepackage[capitalize,noabbrev]{cleveref}

\theoremstyle{plain}

\theoremstyle{definition}

\theoremstyle{remark}

\usepackage[disable,textsize=tiny]{todonotes}
\icmltitlerunning{RadixMLP - Intra-batch Deduplication for Causal Transformers}

\begin{document}

\twocolumn[
% \icmltitle{RadixMLP: Efficient Prefix Deduplication for Batch Transformer Inference}
\icmltitle{RadixMLP - Intra-batch Deduplication for Causal Transformers}

% unless you've provided the [accepted] option to the icml2026



\begin{icmlauthorlist}
\icmlauthor{Michael Feil}{baseten}
\icmlauthor{Julius Lipp}{lipp} 
\end{icmlauthorlist}

\icmlaffiliation{baseten}{Baseten, work done outside of Baseten}
\icmlaffiliation{lipp}{Independent}

\icmlcorrespondingauthor{Michael Feil}{mail@michaelfeil.eu}

\icmlkeywords{Transformer Inference, Batch Processing, Prefix Sharing, Inference Optimization, Machine Learning Systems}

\vskip 0.3in
]



\printAffiliationsAndNotice{}  % leave blank if no need to mention equal contribution

\begingroup
\renewcommand{\thefootnote}{\fnsymbol{footnote}}
\footnotetext[2]{Code available at \url{https://github.com/michaelfeil/radix-mlp}}
\endgroup

\begin{abstract}
Batch inference workloads for causal transformer models frequently process sequences that share common prefixes, such as system prompts, few-shot examples, or shared queries. Standard inference engines treat each sequence independently, redundantly recomputing identical MLP activations for every copy of the shared prefix. We introduce \textbf{RadixMLP}, a technique that exploits the position-wise nature of MLPs, LayerNorms, linear projections, and embeddings to eliminate this redundancy. RadixMLP dynamically maps batches to a prefix trie, gathering shared segments into a compressed representation for position-wise computation and scattering results back only at attention boundaries. RadixMLP is stateless and operates within a single forward pass. In end-to-end serving benchmarks on MS~MARCO v1.1 with Qwen3 models (0.6B to 8B parameters), RadixMLP achieves 1.44--1.59$\times$ speedups in realistic reranking workloads, with up to $5\times$ speedups on synthetic benchmarks with longer shared prefixes. Our code is available at \url{https://github.com/michaelfeil/radix-mlp}.
\end{abstract}

\section{Introduction}
\label{introduction}

Many high-throughput causal transformer deployments process batches whose inputs share long, identical prefixes. Embedding models prepend global context to every document; cross-encoder rerankers pair each candidate passage with the same query; and classification systems reuse the same few-shot examples across inputs. Even in modern serving stacks, such shared prompt structures are common and create substantial computational redundancy.

Standard inference engines nonetheless treat each sequence independently. Even with ragged layouts, duplicate prefix tokens are materialized as distinct rows in GPU memory and the model recomputes identical activations for each copy. This waste is particularly acute during prefill workloads, where \emph{position-wise} components (MLPs, LayerNorms, and linear projections) account for a large fraction of FLOPs.

We introduce \textbf{RadixMLP}, a technique that eliminates this arithmetic inefficiency. RadixMLP relies on a basic property of the Transformer block: while causal Self-Attention is a sequence-mixing operation, the MLP, LayerNorm, linear projections (Q, K, V, O), and embeddings are \emph{position-wise}. For tokens with identical causal history (i.e., the same prefix path), these computations are identical and can be reused.

RadixMLP realizes this reuse by dynamically mapping each batch to a prefix trie. We \emph{gather} shared segments into a compact representation for position-wise computation and \emph{scatter} results back only at the attention boundary. Unlike KV caching approaches that require persistent state management, RadixMLP is entirely stateless and operates within a single forward pass, making it well-suited for batch workloads where cache maintenance is hard to schedule and orchestrate. Additionally, we show that RadixMLP is compatible with autograd, potentially enabling new opportunities for performance improvements in training systems. 

We make the following contributions:

\begin{itemize}
    \item We propose \textbf{RadixMLP}, the \textbf{first stateless approach} to prefix deduplication in Transformer inference, building an alternative to persistent KV caches while achieving compute reuse within a single forward pass.
    \item We show how to integrate RadixMLP into ragged inference by scattering only at the attention operation and gathering back afterward, preserving causal correctness.
    \item We open-source RadixMLP$^\dagger$ with efficient gather/scatter kernels and CPU-side index precomputation, and upstream it into TEI and Candle.
    \item We evaluate on Qwen3 (0.6B--8B) and demonstrate up to $1.4-1.59\times$ speedup on real-world tasks; an Amdahl-style model predicts observed speedups across model sizes.
\end{itemize}

\section{Background}
\label{sec:background}

\subsection{The Causal Transformer Block}
The standard causal (decoder-only) Transformer architecture\citep{vaswani2017attention} 
consists of alternating self-attention and multi-layer perceptron (MLP) blocks, each wrapped 
with residual connections and normalization. Given hidden states $H_l \in \mathbb{R}^{L \times d}$ 
at layer $l$, a typical modern block computes:
\begin{align}
    H'_l &= H_l + \text{Attn}(\text{Norm}(H_l)) \\
    H_{l+1} &= H'_l + \text{MLP}(\text{Norm}(H'_l)) \label{eq:mlp}
\end{align}

These operations differ fundamentally in their data dependencies. The attention block 
contains the attention operation, a \emph{sequence-mixing} operation: under causal masking, the output at position $i$ 
depends on all positions $j \leq i$. In contrast, the MLP, any linear layer, normalization, and 
embedding/projection layers are \emph{position-wise}; they apply the same function 
independently to each token vector. 
By induction across layers, if tokens share identical causal history, their position-wise computations remain identical. Embeddings, LayerNorm, projections, and MLP are per-token operations.
Attention for prefix tokens depends only on earlier identical prefix tokens, thus attention outputs for those tokens are identical, preserving the compact representation for the next layer.

Modern architectures commonly employ gated linear layers such as SwiGLU\citep{shazeer2020glu} in the MLP block. 
For a single token vector $h \in \mathbb{R}^d$:
\begin{equation}
    \text{MLP}(h) = W_{\text{down}} \cdot \bigl( \sigma(W_{\text{gate}} \cdot h) 
    \odot (W_{\text{up}} \cdot h) \bigr),
    \label{eq:swiglu}
\end{equation}
where $W_{\text{up}}, W_{\text{gate}} \in \mathbb{R}^{d_{\text{int}} \times d}$, 
$W_{\text{down}} \in \mathbb{R}^{d \times d_{\text{int}}}$, $\sigma(\cdot)$ is typically 
SiLU, and $\odot$ denotes element-wise multiplication. The intermediate dimension 
$d_{\text{int}}$ is typically several times larger than $d$, making the MLP's three 
matrix multiplications (contributing $\approx 6 d \cdot d_{\text{int}}$ FLOPs per token) 
a major component of prefill compute.

The position-wise nature of Eq.~\ref{eq:swiglu} has a key implication: for any set of 
sequences sharing an identical prefix, the MLP inputs and outputs for that prefix are 
deterministic and identical, regardless of subsequent tokens. This property forms the 
foundation of our approach.

\subsection{Batched Inference and Memory Layouts}
In high-throughput inference serving, multiple requests are batched to maximize GPU utilization. Early implementations used \emph{padding} to align sequences to the longest length in the batch, masking out pad tokens during computation. However, this wastes compute on padding and consumes excessive memory.

Modern engines (e.g., TensorRT-LLM, text-embeddings-inference) utilize a \emph{ragged} memory layout \citep{zeng2022unpadding}. This approach eliminates padding entirely, using variable-length inference where sequences of different lengths are packed contiguously without artificial padding tokens. 
Given a batch of $B$ sequences with lengths $L_1, \dots, L_B$, the tokens are concatenated into a single contiguous tensor $T \in \mathbb{R}^{(\sum L_i) \times d}$. Auxiliary data structures, such as cumulative sequence length arrays (commonly called \texttt{cu\_seqlens}), track the boundaries of individual requests without a batch dimension or materializing the attention mask.

\section{Approach}
\label{sec:approach}
RadixMLP deduplicates position-wise computations for tokens that share identical causal history (i.e., the same prefix path) by compacting the batch via a prefix trie and applying gather/scatter index maps.

\subsection{Problem Formulation}
Consider a batch of $B$ sequences $\mathcal{S} = \{S_1, \dots, S_B\}$, where each sequence $S_i = (t_{i,1}, \dots, t_{i, L_i})$ consists of $L_i$ tokens. In ragged layout of a transformer, \citep{zeng2022unpadding}, these sequences are concatenated into a single flat tensor $X \in \mathbb{R}^{N \times d}$, where $N = \sum L_i$.

A position-wise function $f: \mathbb{R}^d \to \mathbb{R}^{d'}$ (such as an MLP, linear projection, or LayerNorm) is applied row-wise:
\begin{equation}
    Y = f(X) \implies Y_k = f(X_k) \quad \forall k \in [1, N]
\end{equation}
When sequences share a common prefix, the hidden states for the shared prefix positions are identical across those sequences. This property is exploited by many approaches, such as KV caching in SGLang and vLLM \citep{zheng2023sglang,kwon2023pagedattention}. Specifically, if $S_i$ and $S_j$ share a prefix of length $p$, then for all $k \le p$, the hidden states $h_{i,k}$ and $h_{j,k}$ match. This makes calculating $f(h_{i,k})$ and $f(h_{j,k})$ separately computationally wasteful. Importantly, this reuse is only valid for tokens with identical \emph{causal history}; identical $(\text{token\_id}, \text{position\_id})$ pairs appearing after different prefixes must not be merged.

\subsection{The RadixMLP Layout}

To eliminate this redundancy, RadixMLP transforms the batch into a compact representation derived from a prefix trie. We define a mapping between the logical trie nodes and the physical GPU memory layout.

\begin{figure*}[t]
\centering
\begin{tikzpicture}[
    font=\scriptsize,
    box/.style={draw, rounded corners, align=center, inner sep=2pt, minimum height=8mm},
    compact/.style={box, fill=black!6},
    cpu/.style={box, fill=black!10},
    arr/.style={-{Stealth[length=2.1mm,width=2.1mm]}, thick, shorten >=1.5pt, shorten <=1.5pt},
    darr/.style={arr, dashed, opacity=0.85},
    lbl/.style={midway, above, fill=white, rounded corners=1pt, inner sep=2pt},
    node distance=9mm
]
% Main pipeline: compact -> pos-wise -> full(attn) -> compact -> pos-wise -> compact.
\node[compact, text width=0.11\textwidth] (x) {Hidden State\\$N'$};
\node[compact, right=of x, text width=0.14\textwidth] (c1) {Pos-wise ops\\$N'$};
\node[box, right=of c1, text width=0.11\textwidth] (a) {Attention\\$N$};
\node[compact, right=of a, text width=0.14\textwidth] (c2) {Pos-wise ops\\$N'$};
\node[compact, right=of c2, text width=0.11\textwidth] (y) {Hidden State\\$N'$};

\draw[arr] (x) -- (c1);
\draw[arr] (c1) -- node[lbl]{scatter} (a);
\draw[arr] (a) -- node[lbl]{gather} (c2);
\draw[arr] (c2) -- (y);

% CPU scheduler computes indices used by gather/scatter.
\node[cpu, above=7mm of a, text width=0.62\textwidth] (sched) {CPU scheduler: build prefix trie and indices\\$I_{\text{gather}}, I_{\text{scatter}}$ (for next batch)};
\draw[darr] (sched.south west) -- (c1.north);
\draw[darr] (sched.south east) -- (c2.north);
\end{tikzpicture}
\caption{\textbf{RadixMLP integration into causal transformers.} The hidden state remains in compact space ($N'$ tokens) between layers. Position-wise operations (pre-attention norm, projections, MLP) run in compact space. Only the attention operation requires the original space ($N$ tokens), with scatter before and gather after attention (see Algorithm~\ref{alg:radix_mlp_causal_lm}).}
\label{fig:radixmlp_schematic}
\end{figure*}
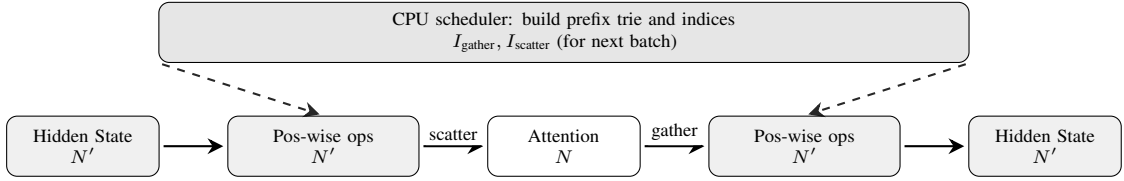

\paragraph{Trie Construction.} 
We construct a prefix trie over the batch. Each trie node is identified by its parent and the next $(\text{token\_id}, \text{position\_id})$ pair, so nodes are path-specific. No GPU-resident tree structure is required; in practice, we only materialize the resulting index maps ($I_{gather}$, $I_{scatter}$) for GPU execution. The root represents the sequence start. For a batch $\mathcal{S}$, we insert every sequence into this trie. Sequences with a shared prefix will traverse the same path; divergent tokens create new branches. Let $\mathcal{U}$ be the set of unique nodes in the trie. The size of the compressed batch is $N' = |\mathcal{U}|$. In batch workloads with shared prefixes, typically $N' \ll N$.  We define the compact-token ratio $\gamma = N'/N \in (0,1]$ and the corresponding compression ratio $r = N/N' = 1/\gamma$. 
% todo this is a duplicate.
Throughout this paper, we consistently use $\gamma$ for the compact-to-original ratio ($N'/N$) and $r$ for the compression ratio ($N/N'$).

\paragraph{Gather and Scatter Kernels.}
To interface with standard PyTorch/CUDA kernels, we linearize the trie into a dense tensor $X' \in \mathbb{R}^{N' \times d}$. We define two index mapping tensors:

\begin{enumerate}
    \item \textbf{Gather Indices ($I_{gather} \in \mathbb{N}^{N'}$):} For each position $i$ in the compact buffer (each unique trie node / prefix-path state), $I_{gather}[i]$ specifies which position in the original input to read. This selects a representative occurrence for that compact token.
    \item \textbf{Scatter Indices ($I_{scatter} \in \mathbb{N}^{N}$):} For each position $j$ in the original layout, $I_{scatter}[j]$ specifies which position in the compact result buffer to read. This broadcasts computed results back to all duplicate positions.
\end{enumerate}

The forward pass for the MLP block is modified as follows:
\begin{align}
    X_{unique} &= X[I_{gather}] \\
    Y_{unique} &= \text{MLP}(X_{unique}) \\
    Y_{restored} &= Y_{unique}[I_{scatter}]
\end{align}
Since the MLP has complexity $\mathcal{O}(N d^2)$, the cost is reduced to $\mathcal{O}(N' d^2)$. As the gather/scatter operations are memory-bound $\mathcal{O}(N)$ copies, the net speedup approaches the compression ratio $r = N/N'$ as the hidden dimension $d$ grows.
By construction, this transformation works for arbitrary trie patterns, with granularity of branching at the token level.
To realize this speedup in practice, we implement efficient copy kernels$^\dagger$. Using fully reproducible benchmarks in our repository, we achieve 2$\times$--4$\times$ speedup over PyTorch 2.1 and 3$\times$--22$\times$ speedup over Candle 0.9.2 on NVIDIA H100 GPUs across various tensor shapes (see \cref{sec:appendix:scatter_gather_benchmarks} for details).

\subsection{Zero-Overhead Scheduling}
Constructing the trie and generating index maps on the GPU would introduce synchronization overheads that could negate the compute savings. To mitigate this, we utilize the concept of a \emph{pipelined scheduler}, first pioneered by \citet{zhu2025nanoflowoptimallargelanguage}.

In most inference systems, the scheduling loop operates asynchronously on the CPU. While the GPU executes batch $t$, the CPU scheduler analyzes the request queue for batch $t+1$. We extend the scheduler to: (1) Construct the prefix trie; and (2) Pre-calculate $I_{gather}$ and $I_{scatter}$. 
Due to an efficient Rust implementation, computing approximately 16,384 tokens takes between $129\,\mu\mathrm{s}$ to $750\,\mu\mathrm{s}$ on a single-thread Intel Xeon CPU depending on the prefix ratio (see~\cref{tab:radix_mlp_performance}).
This is around 3 orders of magnitude below the corresponding GPU inference time of the according model, making even synchronous usage feasible. 

\subsection{Integration with Attention}
A critical requirement for RadixMLP is \emph{causal consistency}. While position-wise layers are independent per token, the Attention mechanism requires the full context window.

To maintain correctness while minimizing memory operations, we place \emph{Scatter} and \emph{Gather} operations at the boundary between position-wise and sequence-mixing computation. 
Concretely, we compute all position-wise operations in compact space: the pre-attention LayerNorm, RoPE, Embedding Layer, Q/K norms (for models like Qwen3), and the Q, K, V projections on $N'$ tokens.
Only the attention operation requires the original full layout. The residual addition across layers is also position-wise and thus runs in compact space, reducing activation memory of this buffer by a factor of $r$.
We then \emph{scatter} the resulting Q/K/V tensors back to the full $N$-token ragged layout and run the attention operation (FlashAttention) in full space. 
Immediately after attention operation, we \emph{gather} the attention output back to compact space for the O projection, post-attention LayerNorm, MLP, and residual additions.

Position indices are preserved alongside compact tokens: we gather the position ID tensor using $I_{gather}$, so each compact token retains its original sequence position for RoPE computation. Since scattering restores the original ragged order, standard attention metadata (e.g. \texttt{cu\_seqlens}) remain unchanged. The complete pseudocode for integrating RadixMLP into a causal language model forward pass is provided in~\cref{alg:radix_mlp_causal_lm}.

\subsection{RadixMLP for the backward pass}
While the focus of this work is inference, RadixMLP is compatible with training mode because the compaction operators are differentiable. Concretely, the compacting gather is an \texttt{index\_select} and its backward pass is a scatter-add into the original layout (\texttt{index\_add}), accumulating gradients when multiple original tokens map to the same compact representative. This is the same mechanism commonly used in message passing and pooling operators in geometric deep learning.

\paragraph{IndexSelect forward and backward.}
Let $y=\mathrm{IndexSelect}(x,I)$, i.e.
\[
y_j = x_{I_j}.
\]
For a loss $\mathcal{L}(y)$, the backward pass is
\[
\frac{\partial \mathcal{L}}{\partial x_k}
= \sum_{j:\, I_j = k} \frac{\partial \mathcal{L}}{\partial y_j},
\]
i.e., gradients are scattered back to $x$ and summed when indices repeat (\texttt{index\_add}). Therefore, RadixMLP's gather/scatter operation is fully compatible with autograd.

\paragraph{Validation and a note on attention kernels.}
We validated both forward and backward correctness on a 2-layer Qwen3-style model (hidden size 256, intermediate size 512) across 
synthetic prefix-sharing patterns (single sequence, identical sequences, shared prefix, no sharing, mixed lengths, complex sharing) (see \cref{sec:appendix:training} for detailed results). 
With PyTorch \texttt{SDPA} attention as the reference implementation, enabling RadixMLP produces numerically identical logits up to tolerance (\texttt{rtol}=1e\text{-}4, \texttt{atol}=1e\text{-}4) and gradient differences on the order of $10^{-5}$ or below across all test cases (maximum observed $\approx 1.9\times 10^{-5}$).

During early experiments we observed substantially larger backward differences (up to $3\times 10^{-2}$) when comparing against FlashAttention-2 (\texttt{FA2}).
Importantly, these discrepancies were \emph{not} caused by RadixMLP: the same magnitude of differences appeared even with RadixMLP disabled. 
In other words, \emph{swapping the attention implementation alone} can introduce larger forward and backward numerical variation than toggling RadixMLP on/off. 
Switching from fp16 to fp32 improved the numerical correctness by two orders of magnitude. 
In our ablation study, the forward-pass maximum logit difference between FA2 and SDPA was $2.2\times 10^{-4}$ regardless of RadixMLP. 
The corresponding backward-pass maximum gradient difference was $1.3\times 10^{-2}$ to $3.6\times 10^{-2}$, again largely independent of RadixMLP.

These results support two conclusions: (i) RadixMLP is gradient-correct for the position-wise components it modifies, and (ii) in practical training pipelines, observed backward differences are often dominated by the choice of precision and various other kernels such as attention operation.
Their numerical behavior dominates the batch-variant effects caused by RadixMLP's compaction itself. We include a fully reproducible test suite $^\dagger$ that evaluates multiple sequence shapes and attention backends and reports both forward and backward differences.

\paragraph{Takeaway.}
RadixMLP's gather/scatter operations preserve differentiability and yield gradients that match the non-radix baseline up to typical floating-point tolerances. When larger discrepancies occur in end-to-end comparisons, they are more plausibly attributable to attention-kernel numerical differences than to RadixMLP. 
We leave further experimentation with the effects on large training runs to future work and focus on gains during inference.

\section{Experiments}
\label{sec:experiments}

We evaluate RadixMLP with two benchmarks: (i) synthetic forward-pass microbenchmarks on controlled prefix/suffix batches, and (ii) a real-world end-to-end serving benchmark on MS~MARCO v1.1 query-passage pairs \citep{bajaj2016msmarco} and (iii) compare it against other inference frameworks.

\subsection{Experimental Setup}
\label{subsec:experimental-setup}

All experiments run on an NVIDIA H100 (80GB) GPU with FlashAttention-2 \citep{dao2023flashattention2}. We evaluate three variants of the Qwen3 embedding models \citep{yang2025qwen3}: \texttt{Qwen3-0.6B}, \texttt{Qwen3-4B} and \texttt{Qwen3-8B}. We run all models in float16 (fp16) precision. 
For end-to-end measurements, we use text-embeddings-inference (TEI) \citep{dehane2023textembeddingsinference}, with candle-cuda backend and vllm 0.13.0 with block size of 32. 

We present three different inference benchmarks:
\begin{itemize}
    \item (a) A synthetic, fixed batch size of $B=32$ sequences, varying the \emph{prefix length} (shared across all sequences) and \emph{suffix length} (unique per sequence). Short prefixes (32--256 tokens) model query-style sharing common in reranking; longer prefixes (512--2048 tokens) model instruction-style sharing common in few-shot classification. This gives us a uniform baseline, to exactly profile the forward pass time of each configuration.
    \item (b) A constructed MS~MARCO v1.1 dataset \citep{bajaj2016msmarco}. We use the validation split. For each row in the dataset, we pair each query with all passage options into the Qwen3Reranker template. This dataset has a distribution of 75 - 200 tokens per sequence. Empirically we validated that when using a block-size of 32, around 32 \% can be reused with a KV cache implementation. The template formats each query-passage pair with a shared system prompt, creating natural prefix sharing across passages for the same query.
    \item (c) An augmentation of the MS~Marco v1.1 dataset \citep{bajaj2016msmarco} by using the larger train split. We flip the document with the query, and shorten the system prompt by removing the default instruction. This dataset is shortened to 65-200 tokens. Empirically we validated the re-usable block-size of 32, which results in only 20\% of tokens being reusable with a blocked KV cache implementation. The flipped order reduces prefix sharing since queries become the unique suffix rather than the shared prefix.
\end{itemize}

Specifically we have chosen to construct these datasets, as evaluating models on MS~MARCO \citep{bajaj2016msmarco} is a subtask in many benchmarks such as MTEB \citep{muennighoff2023mtebmassivetextembedding}. Besides the application in commercial inference systems, we want to facilitate a faster evaluation of models on these benchmarks when testing a new set of weights. 

\subsection{Synthetic dataset results}

For the synthetic dataset in \cref{subsec:experimental-setup}, we are evaluating an isolated forward pass of the model. 
Using the candle-cuda backend for all tasks, we vary the prefix and suffix lengths, the qwen3 variant, and enabling or disabling RadixMLP.

\begin{figure}[t]
\centering
\includegraphics[width=\columnwidth]{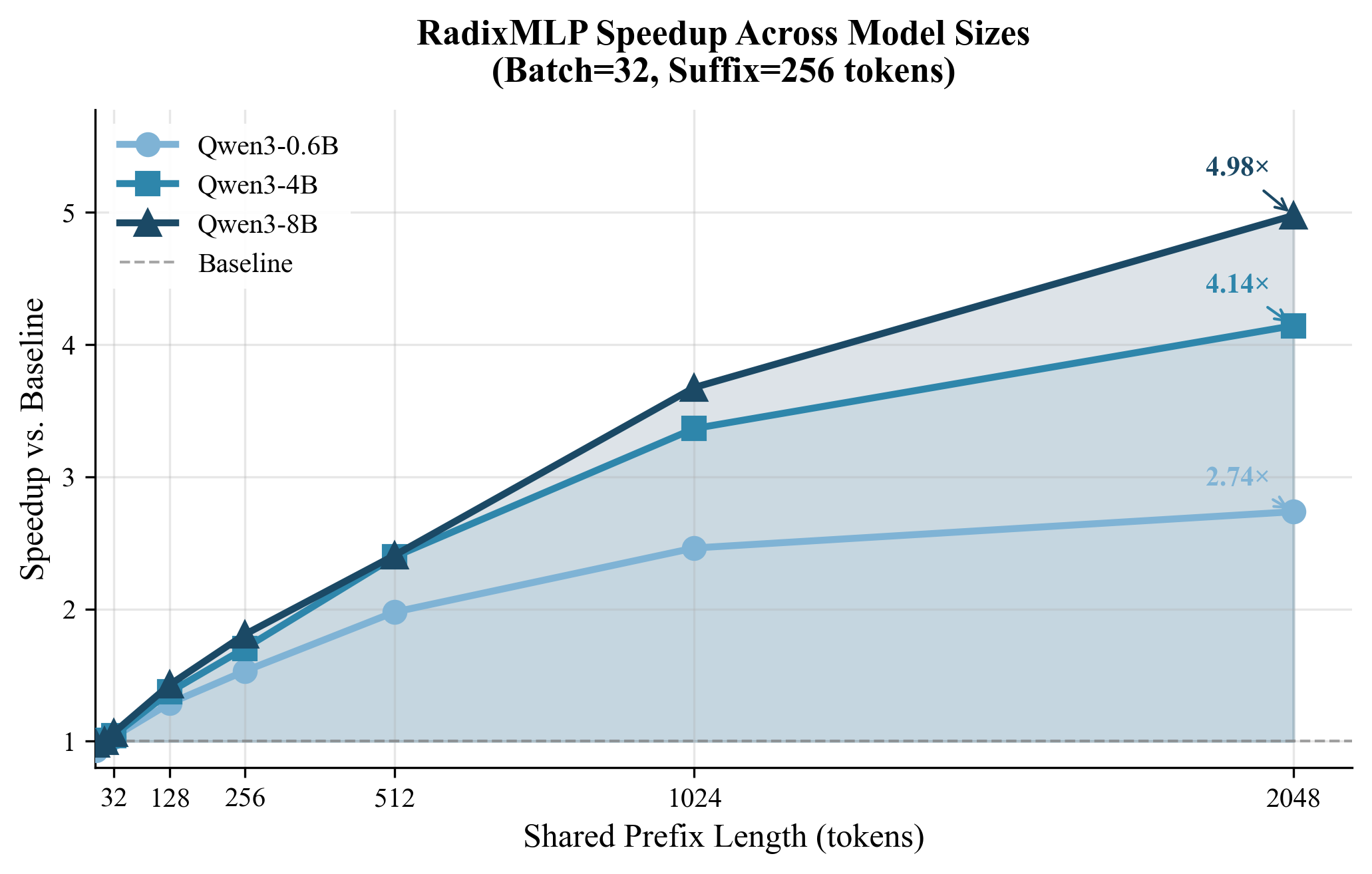}
\caption{\textbf{RadixMLP speedup over prefix lengths with fixed suffix length.} Larger models benefit more as position-wise operations dominate the FLOP budget. Qwen3-8B achieves 5.0$\times$ speedup at 2048-token prefixes. Figure~\ref{fig:radixmlp_suffix_comparison} shows additional configurations. }
\label{fig:radixmlp_comparison}
\end{figure}

Figure~\ref{fig:radixmlp_comparison} compares the forward pass latency with RadixMLP enabled versus the baseline across prefix/suffix configurations. As described in Section~\ref{sec:approach}, trie construction and index computation are performed asynchronously by the CPU scheduler; all timings reflect pure GPU execution.

The results demonstrate three key trends:
\begin{enumerate}
    \item \textbf{Prefix length}: Longer shared prefixes yield lower $\gamma$ (higher $r$) and greater speedups. The 2048/256 configuration achieves up to 5.0$\times$ speedup on Qwen3-8B.
    \item \textbf{Model size}: Larger models benefit more from RadixMLP, as MLP compute dominates over gather/scatter overhead.
    \item \textbf{Suffix ratio}: Shorter suffixes (higher prefix:suffix ratio) maximize compression benefit.
\end{enumerate}
Table~\ref{tab:amdahl} additionally reports absolute forward pass latencies for the 2048/256 configuration.

\subsection{End-to-end results}
\label{sec:end-to-end-results}

We evaluate a real serving pipeline by benchmarking TEI \citep{dehane2023textembeddingsinference} on MS~MARCO v1.1 validation query--passage pairs \citep{bajaj2016msmarco}. 

Each request embeds multiple query--document pairs formatted with a reranker-style Qwen3 chat template. This end-to-end measurement includes tokenization, dynamic batching, and HTTP overhead in addition to the model forward pass.

We compare TEI with RadixMLP disabled against TEI with RadixMLP enabled. We set \texttt{radix\_mlp\_threshold} to 0.95 (enabled) and compare against 0.0 (disabled), enabling RadixMLP only when at least 5\% of tokens in a batch can be deduplicated. We report results at \texttt{max\_batch\_tokens=65536}; results are similar at 16384.

We report per-request latency over 1287 requests (median and mean). We provide additional details in \cref{sec:appendix:end-to-end-results}.

\begin{table}[h]
\centering
\small
\caption{End-to-end request latency (seconds) in TEI on MS~MARCO v1.1 validation (1287 requests), with RadixMLP enabled vs.\ disabled, at \texttt{max\_batch\_tokens=65536}.}
\label{tab:tei_end_to_end}
\begin{tabular}{@{}lccccc@{}}
\toprule
 & \multicolumn{2}{c}{P50 (s)} & \multicolumn{2}{c}{Mean (s)} & \\
\cmidrule(lr){2-3} \cmidrule(lr){4-5}
Model & Base & Radix & Base & Radix & Speedup \\
\midrule
Qwen3-0.6B & 0.78 & 0.54 & 0.78 & 0.54 & 1.44$\times$ \\
Qwen3-4B   & 3.76 & 2.42 & 3.75 & 2.41 & 1.56$\times$ \\
Qwen3-8B   & 5.96 & 3.74 & 5.96 & 3.75 & 1.59$\times$ \\
\bottomrule
\end{tabular}
\end{table}

Table~\ref{tab:tei_end_to_end} shows that RadixMLP reduces mean latency by 1.44$\times$ (0.6B), 1.56$\times$ (4B), and 1.59$\times$ (8B), with similar improvements at the median. These gains are smaller than the synthetic forward-pass speedups because prefix sharing is less extreme in this workload and because end-to-end latency includes non-model overheads, but the improvement remains substantial in a practical serving setting.

\paragraph{Understanding speedup factors}

We explain the observed speedups with a simple Amdahl-style model that separates position-wise computation (executed in compact space) from the attention operation (executed in full space); Table~\ref{tab:amdahl} shows close agreement with measurements.

As described in Section~\ref{sec:approach}, RadixMLP runs all position-wise components on the compacted $N'$ tokens, and expands only for the attention operation on the full $N$ tokens. Let $f_c$ denote the fraction of FLOPs attributable to position-wise work. A simple per-token FLOP proxy gives
\[
f_c = \frac{8d^2 + 6d\,d_{\text{int}}}{8d^2 + 6d\,d_{\text{int}} + 4Ld},
\]
where $d$ is the hidden size, $d_{\text{int}}$ is the MLP intermediate size, and $L$ is the sequence length.\footnote{LayerNorm and RoPE contribute lower-order terms; including them increases $f_c$ slightly.}
Because projections and MLP scale as $\mathcal{O}(d^2)$ while attention scales as $\mathcal{O}(Ld)$, $f_c$ increases with model size (for $L{=}2304$: $f_c \approx 73\%$, $88\%$, $92\%$ for Qwen3-0.6B/4B/8B).

Let $r$ denote the compression ratio defined in Section~\ref{sec:approach}. Ignoring small gather/scatter overheads, the expected speedup is
\[
\text{Speedup} \;=\; \frac{1}{(1-f_c) + f_c/r}.
\]

\begin{table}[h]
\centering
\small
\caption{Theoretical model vs. observed speedups for 2048/256 configuration, with absolute forward pass latencies (ms).}
\label{tab:amdahl}
\begin{tabular}{@{}lccccc@{}}
\toprule
 & & \multicolumn{2}{c}{Speedup} & Latency (ms) \\
\cmidrule(lr){3-4}
Model & $f_c$ & Predicted & Observed & Base $\to$ Radix \\
\midrule
Qwen3-0.6B & 73\% & 2.68$\times$ & 2.74$\times$ & 394 $\to$ 144 \\
Qwen3-4B   & 88\% & 4.14$\times$ & 4.14$\times$ & 1656 $\to$ 400 \\
Qwen3-8B   & 92\% & 4.82$\times$ & 4.98$\times$ & 2866 $\to$ 576 \\
\bottomrule
\end{tabular}
\end{table}

A slight excess speedup for Qwen3-8B ($\sim$3\%) may reflect secondary effects such as reduced activation memory traffic and improved cache locality.
\paragraph{Crossover Analysis}

A key practical question is: \emph{when does RadixMLP provide benefit?} RadixMLP trades additional overhead (gather/scatter memory traffic and index bookkeeping) for reduced position-wise FLOPs. When redundancy is low ($\gamma \approx 1$), the saved compute is too small to amortize this overhead, motivating a crossover point and a simple gating threshold.

In our setup, the crossover point where speedup exceeds 1.0$\times$ occurs at short prefix lengths (on the order of tens of tokens) for typical suffix configurations (as discussed in \cref{sec:discussion}). For practical batch workloads with prefixes $\geq$32 tokens, RadixMLP consistently provides benefit over adding small copy kernels for gather and scatter operations.

\subsection{End-to-end comparison against vLLM}
Selecting two configurations from \cref{sec:end-to-end-results} with 16384 max-batch-tokens setting, we are comparing TEI against vLLM v0.13.0. vLLM uses a block-size of 32 tokens.

As datasets we use configurations (b) and (c) from \cref{subsec:experimental-setup}. 
While dataset (b) has some advantages for intra batch reuse as it's prefixed with the query, giving room for approximately 10 extra tokens that can be reused if identical queries land in the same batch. 
Dataset (c) has a slightly shorter system prompt, and has fewer opportunities for cache usage overall, either in a block-based or intra-batch deduplication.

The results show that TEI without RadixMLP is the worst performing inference system (~\cref{tab:vllm_vs_tei}). With RadixMLP enabled, TEI outperforms vLLM for the Qwen3-0.6B model (0.55s vs 0.71s p50). 
For bigger Qwen3-4B and Qwen3-8B models in~\cref{tab:vllm_vs_tei}, vLLM outperforms TEI by a 3-7 \% margin in p50.

Overall, this shows that a collection of relatively simple kernels in TEI can be competitive with more sophisticated systems like vLLM. 
Beyond the comparison in the table, a key difference is memory usage. In TEI, no additional memory is allocated for a paged KV cache, leaving a memory footprint of $\approx 17GB$ vs $\approx 79GB$ for Qwen3-8B.

\begin{table}[h]
\label{tab:vllm_vs_tei}
\centering
\small
\caption{Comparison of TEI with and without RadixMLP against vLLM. p50 response time in s (p90 response time in s),  Lower is better. Dataset (b) with a ratio of 0.61, and dataset (c) with a ratio of 0.71.}
\begin{tabular}{@{}lcccc@{}}
\toprule
Model &  $\gamma$ & TEI & TEI RadixMLP & vLLM \\
\midrule
Qwen3-0.6B & 0.61 & 0.78 (0.85) & \textbf{0.55 (0.60)} & 0.71 (0.79) \\
Qwen3-0.6B & 0.71 & 0.70 (0.77) & \textbf{0.57 (0.63)} & 0.71 (0.83) \\
Qwen3-4B & 0.61 & 3.65 (3.98) & 2.39 (2.62) & \textbf{2.22 (2.51)} \\
Qwen3-4B & 0.71 & 3.33 (3.66) & 2.52 (2.78) & \textbf{2.44 (2.53)} \\
Qwen3-8B & 0.61 & 5.75 (6.27) & 3.66 (4.00) & \textbf{3.40 (3.84)} \\
Qwen3-8B & 0.71 & 5.23 (5.75) & 3.88 (4.28) & \textbf{3.77 (3.91)} \\
\bottomrule
\end{tabular}
\end{table}

\section{Discussion}
\label{sec:discussion}

\subsection{Stateless Design}
Unlike PagedAttention \citep{kwon2023pagedattention} or RadixAttention \citep{zheng2023sglang}, which require persistent KV caches with block tables, eviction policies, and distributed coordination, RadixMLP operates entirely within a single forward pass. This makes it valuable for batch inference where maintaining caches for millions of heterogeneous documents is impractical, and enables libraries like TEI \citep{dehane2023textembeddingsinference} to achieve cache-like speedups without cache management overhead. We contributed RadixMLP to TEI's overlap scheduler for integration in the 2026 major release, enabling prefix reuse without KV cache block granularity and allowing near-identical sequences to be processed efficiently regardless of block size constraints.

\subsection{Limitations}
\paragraph{Low-redundancy batches.}
Batches with minimal redundancy ($r \approx 1$, i.e., $N'/N \approx 1$) incur small overhead from index computation; our implementation bypasses RadixMLP when the compact-token ratio $\gamma = N'/N$ is above a configurable threshold.

\paragraph{Autoregressive generation.}
For autoregressive generation, RadixMLP benefits are limited to context-phase prompt processing, where KV caching remains more suitable. It additionally could be useful for speculative tree decoding, where duplicated tokens only need to be computed once. 

\paragraph{Numerical differences without batch-invariant kernels.}
RadixMLP's position-wise operations (MLPs, LayerNorms, projections, RoPE) produce bit-identical results regardless of batch organization; RadixMLP has similar lossless guarantees to KV caching techniques. However, without the use of batch-invariant kernels \citep{he2025nondeterminism}, operations such as matmul will have a slight output variance, depending on the input shapes. The \texttt{index\_select} kernels we provided are launched over each token entry and therefore batch-invariant.

\paragraph{Training.}
While we provide proof that RadixMLP's approach of differentiable gather/scatter operations work for training, we leave a detailed training evaluation to future work.

\paragraph{Memory Overhead.}
The gather and scatter index tensors $I_{gather} \in \mathbb{N}^{N'}$ and $I_{scatter} \in \mathbb{N}^{N}$ require $\mathcal{O}(N)$ integers of additional storage. For typical 32-bit indices, this amounts to $4N$ bytes---negligible compared to activation memory ($N \times d \times 2$ bytes for bf16/fp16) when $d \gg 4$.

\paragraph{Long-Context Workloads.}
RadixMLP's benefits diminish as sequence length increases. For very long sequences (32K+ tokens), attention becomes the dominant cost since it scales as $\mathcal{O}(L^2)$ while position-wise operations scale as $\mathcal{O}(L)$. At such lengths, even with high compression ratios, the fraction $f_c$ of FLOPs in compact space decreases substantially, reducing the achievable speedup.

\section{Related Work}
\label{sec:related}
\paragraph{Prefix-Tree based Attention Kernels.}
FlashAttention\citep{dao2022flashattention} and FlashInfer\citep{ye2025flashinferefficientcustomizableattention} accelerate attention with memory-efficient tiling. HydraGen\citep{juravsky2024hydragen} leverages shared prefixes to batch $Q K^\top$ for common contexts, bypassing redundant access. While both HydraGen and RadixMLP exploit prefix sharing, HydraGen speeds up \emph{attention operation} (sequence-mixing: $Q K^\top$, softmax$\cdot V$), whereas RadixMLP targets \emph{position-wise} layers (MLPs, LayerNorms, projections), i.e., per-token computation. These are fully complementary---HydraGen reduces attention cost, RadixMLP reduces cost for the components that dominate prefill FLOPs for short sequences (73--92\% by model size).

\paragraph{Batching and Scheduling.}
Ragged batching\citep{zeng2022unpadding} and prefix-aware scheduling\citep{zheng2025batchllmoptimizinglargebatched} address KV aware group scheduling using the prefix tree, but do not exploit actual input duplication and also rely on a block-based KV cache. 

RadixMLP operates orthogonally: it reduces arithmetic in position-wise layers, and can stack with KV cache, attention optimizations (e.g., HydraGen), or batching/scheduling optimizations.

\paragraph{KV cache memory management and sharing.}
A large line of work accelerates inference by improving how the KV cache is stored, indexed, and reused.
PagedAttention and vLLM\citep{kwon2023pagedattention} introduce paged, block-based KV management with block-level sharing, while SGLang's RadixAttention\citep{zheng2023sglang} organizes prefixes in a radix tree for cross-request KV reuse. These stateful approaches require complex scheduling and are limited by block granularity, whereas RadixMLP offers block-free, stateless reuse within individual batches.

\paragraph{Prefix-sharing attention kernels and cascade inference.}
Orthogonal to KV management, various works in the attention kernel space have utilized the causal prefix structure to get additional gains. 
Beyond FlashAttention series\citep{dao2023flashattention2} as foundation for further improvements, attention kernels have exploited shared prefixes for efficiency gains.
For long-context and multi-level KV settings, FlashInfer's \emph{cascade inference / cascade attention} framework computes attention over shared prefix segments once, 
then merges partial attention states with per-request suffix attention using hierarchical KV layouts\citep{ye2025flashinferefficientcustomizableattention}. \citep{yao2025deftdecodingflashtreeattention} introduces Decoding with Flash Tree-Attention, which reuses previously loaded KV memory if it's shared among many sequences. FlashInfer's Cascade inference de facto increases the number of stages over HydraGen and has to build a Radix Tree for attention, similar to RadixMLP's Tree.

\paragraph{Prefix-aware batching and scheduling.}
Beyond kernels, several systems explicitly \emph{reorder and group requests} to maximize prompt-prefix locality.
BatchLLM\citep{zheng2025batchllmoptimizinglargebatched} identifies common prefixes globally and schedules requests sharing prefixes together, improving KV reuse and overall throughput in large-batch settings; it also combines prefill and decode tokens to better saturate GPUs.
Preble\citep{srivatsa2024preble} extends prefix-aware scheduling to distributed serving, co-optimizing KV reuse and load balancing across multiple GPUs with hierarchical scheduling.
These approaches are complementary to ragged batching\citep{zeng2022unpadding}, which removes padding waste but does not by itself exploit \emph{duplicate} computation arising from identical prefixes. However, both do not distinguish between a sequence that is queued for prefill and that has completed the prefill stage.

\paragraph{Prefix-aware scheduling for training}
Stateful KV cache is not compatible with gradient propagation, and caching would be short-lived during the training stage. Attention kernels, such as FlashInfer, are primarily noted for their technical complexity\citep{ye2025flashinferefficientcustomizableattention}. To our knowledge, no kernel work has explored a backward-compatible cascade-based kernel. Given the missing foundations in training-compatible systems optimization, we found no relevant work on scheduling optimization for more efficient training. We hope to have given the research community some new levers to explore this field. This includes the partitioning of datasets in scheduling multi-node training workloads, and effects of RadixMLP-inspired operations on future training runs.

\section{Future Work}
We identify two promising directions for extending this work:

\paragraph{Post-Training Framework Integration.} Shared prefixes are prevalent in prompt templates used for supervised fine-tuning and reinforcement learning from human feedback. Since the \texttt{index\_select} operations are differentiable with respect to the input tensors, RadixMLP indices can be precomputed in the dataloader. This could potentially enable training optimizations, as padding-based batching introduces substantial inefficiency. The primary challenge lies in developing and achieving adoption for training frameworks that support ragged tensor layouts.

\paragraph{Cascade-Free Attention Kernels.} The precomputed prefix trie could enable attention kernels that process shared prefixes without redundant memory access. Persistent kernels that transform the RadixMLP trie into a block-aligned layout represent a theoretically promising approach, though achieving practical speedups on modern GPUs remains an engineering challenge due to the complexity of scheduling sequential memory access patterns.

\section*{Impact Statement}
RadixMLP reduces energy consumption in batch Transformer inference by eliminating duplicate computation for shared prefixes. Our experiments demonstrate 1.4--1.6$\times$ throughput improvements on realistic reranking workloads, with better improvements on high-prefix-sharing scenarios. At the scale of modern embedding and reranking services processing billions of requests daily, these per-request savings compound into meaningful reductions in datacenter energy consumption and associated carbon emissions.
Beyond this, we hope for broad adoption across other domains to lower the footprint of machine learning systems. 

\bibliography{radixmlp}
\bibliographystyle{icml2026}

% APPENDIX

\newpage
\appendix
\onecolumn
\section{Additional Experimental Results}
\label{sec:appendix}

\subsection{Impact of Suffix Length on RadixMLP Speedup}

\begin{figure}[h]
\centering
\includegraphics[width=0.85\textwidth]{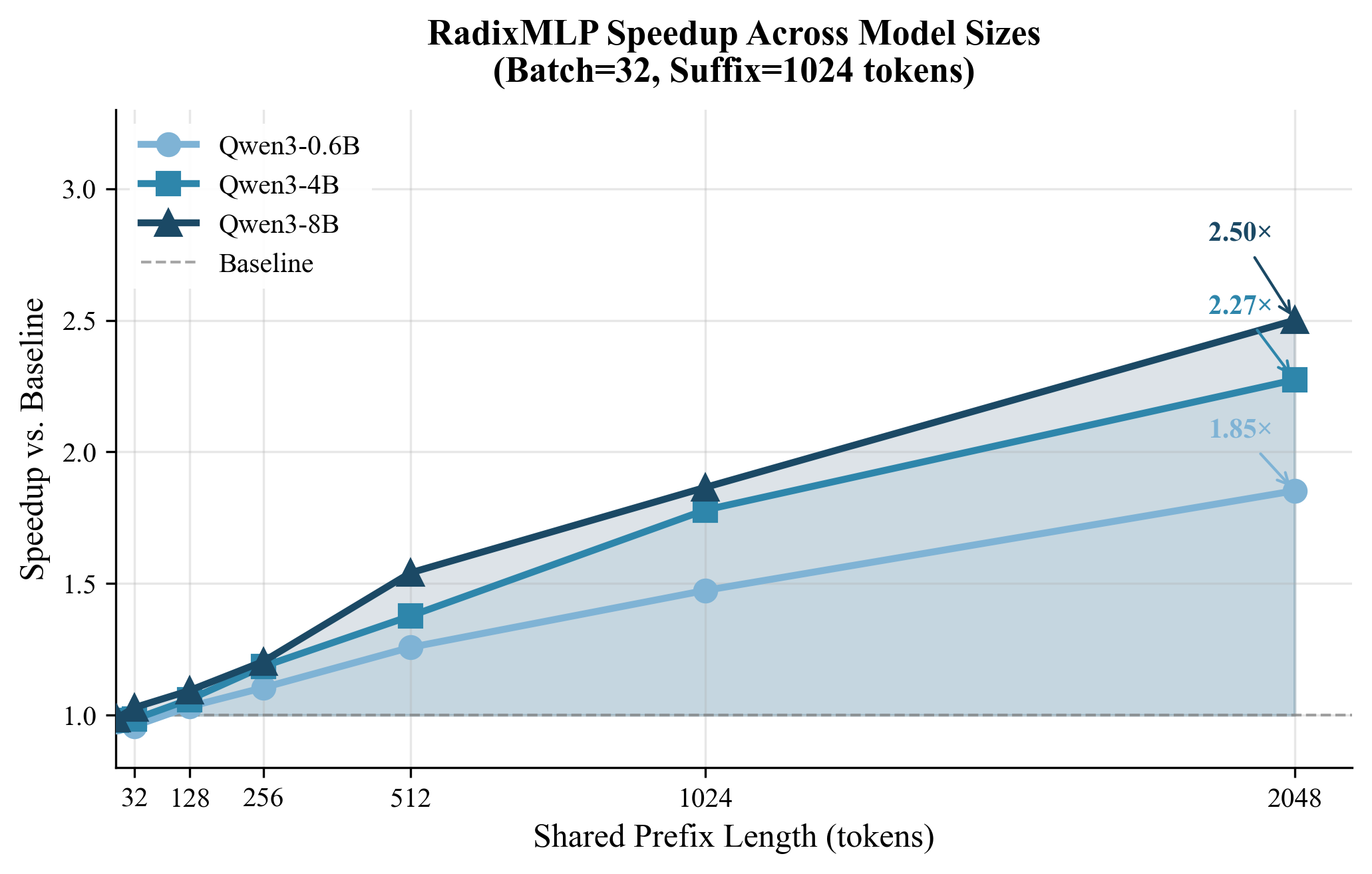}
\caption{\textbf{RadixMLP speedup with Suffix=1024 tokens.} With longer suffixes, the compression ratio decreases, resulting in reduced speedups. Qwen3-8B achieves 2.50$\times$ speedup at 2048-token prefixes. \cref{tab:qwen3-4b-radixmlp-forward} provides the raw numbers for Qwen3-4B.}
\label{fig:radixmlp_suffix_comparison}
\end{figure}

Complementary to \cref{fig:radixmlp_comparison,fig:radixmlp_suffix_comparison} presents RadixMLP speedups across different suffix lengths, with a fixed suffix length of 1024 tokens.
While our main results focus on suffix=256 tokens (representative of embedding and reranking workloads), we also evaluated suffix=1024 tokens to understand how longer unique sequences affect the compression benefit.

% TODO(michaelfeil): We could add another (very long experiment going up all the way to 32k context and seeing the benefit drop again due to the attention portion.

The results confirm that RadixMLP's benefit scales with the \emph{prefix-to-suffix ratio} via the compression ratio $r$ (Section~\ref{sec:approach}). For a batch of $B$ sequences with shared prefix length $P$ and per-sequence suffix length $S$,
\[
r = \frac{N}{N'} = \frac{B(P + S)}{P + BS}.
\]

For the 2048/256 configuration: $r = \frac{32 \times 2304}{2048 + 32 \times 256} = 7.2\times$. For the 2048/1024 configuration: $r = \frac{32 \times 3072}{2048 + 32 \times 1024} = 2.8\times$. This lower compression ratio directly explains the reduced speedup observed with longer suffixes.

Looking at the overhead caused by Qwen-3-4B for small-prefixes (1,1024). Estimating the overhead of running the additional index-select operation at $\approx 40 \mu s$ (8224 -> 8256 tokens) for each two operations times 36 layers, leaves us at $\approx 2.88 ms$.

\begin{table}[H]
\centering
\small
\begin{tabular}{rrrrrrrr}
\toprule
Prefix & Suffix & Batch & Compact & RadixMLP & RadixMLP & Speedup & $\gamma$ \\
& & Tokens ($N$) & Tokens ($N'$) & Enabled (ms) & Disabled (ms) & & \\
\midrule
1    & 256  & 8{,}224  & 8{,}193  & 188.4 & 184.0 & 0.98 & 0.996 \\
16   & 256  & 8{,}704  & 8{,}208  & 190.5 & 191.6 & 1.01 & 0.943 \\
32   & 256  & 9{,}216  & 8{,}224  & 192.4 & 200.9 & 1.04 & 0.892 \\
128  & 256  & 12{,}288 & 8{,}320  & 190.5 & 261.6 & 1.37 & 0.677 \\
256  & 256  & 16{,}384 & 8{,}448  & 202.4 & 344.8 & 1.70 & 0.515 \\
512  & 256  & 24{,}576 & 8{,}704  & 218.6 & 523.7 & 2.40 & 0.354 \\
1024 & 256  & 40{,}960 & 9{,}216  & 263.9 & 888.0 & 3.36 & 0.225 \\
2048 & 256  & 73{,}728 & 10{,}240 & 399.7 & 1655.8 & 4.14 & 0.139 \\
\midrule
1    & 1024 & 32{,}800 & 32{,}769 & 742.2 & 727.2 & 0.98 & 0.999 \\
32   & 1024 & 33{,}792 & 32{,}800 & 743.9 & 732.0 & 0.98 & 0.971 \\
128  & 1024 & 36{,}864 & 32{,}896 & 748.0 & 792.4 & 1.06 & 0.892 \\
256  & 1024 & 40{,}960 & 33{,}024 & 747.5 & 885.5 & 1.18 & 0.806 \\
512  & 1024 & 49{,}152 & 33{,}280 & 780.2 & 1074.0 & 1.38 & 0.677 \\
1024 & 1024 & 65{,}536 & 33{,}792 & 839.8 & 1494.0 & 1.78 & 0.516 \\
2048 & 1024 & 98{,}304 & 34{,}816 & 1019.4 & 2318.1 & 2.27 & 0.354 \\
\bottomrule
\end{tabular}
\caption{
Forward-pass latency for Qwen-3-4B with and without RadixMLP.
All measurements latencies are median wall-clock times under multiple runs with cargo bench.
Speedup is computed as the ratio of RadixMLP-disabled to RadixMLP-enabled latency.
}
\label{tab:qwen3-4b-radixmlp-forward}
\end{table}

\section{Implementation Details}
\label{sec:appendix_impl}

\subsection{Implementation: Integrating RadixMLP into Causal Transformers}
RadixMLP emerged from efforts to implement efficient KV caching, where we observed that computational blocks could be deduplicated.
The technique is most straightforward to implement with ragged layouts, which naturally support the required index operations. In padded formats, implementation remains feasible but requires more complex reshape and index-select operations, making the approach less apparent. 
In ragged layout, RadixMLP can be easily integrated into a causal transformer. 
By adding scatter and gather operations immediately before and after each attention operation, as in ~\cref{alg:radix_mlp_causal_lm}. 
In essence, most operations within the transformer are position-wise. In practice, we can even run the embedding, and positional encoding in compact space, since only the attention requires the original layout.
Each layer performs one scatter to expand compact activations for attention and one gather to return to compact space afterward. 
Concretely, we use $\text{Gather}(\mathbf{t}^{\text{orig}}, \mathbf{I}_{\text{gather}})$ to select representatives into compact space via $\mathbf{t}[j] = \mathbf{t}^{\text{orig}}[\mathbf{I}_{\text{gather}}[j]]$, and $\text{Scatter}(\mathbf{t}, \mathbf{I}_{\text{scatter}})$ to expand back via $\mathbf{t}^{\text{orig}}[i] = \mathbf{t}[\mathbf{I}_{\text{scatter}}[i]]$.

\begin{algorithm}[H]
\caption{RadixMLP-Enhanced Causal Language Model Forward Pass}
\label{alg:radix_mlp_causal_lm}
\begin{algorithmic}[1]
\REQUIRE Input tokens $\mathbf{x} = [x_1, \dots, x_n]$, position IDs $\mathbf{p} = [p_1, \dots, p_n]$, cumulative sequence lengths $\mathbf{s}$, number of layers $N_{\text{layers}}$
\ENSURE Logits $\mathbf{y} \in \mathbb{R}^{n \times |\mathcal{V}|}$

\STATE \textbf{Token Compaction:}
\STATE Build a prefix trie over the batch (path-specific nodes)
\STATE Build compaction indices:
\STATE \quad $\mathbf{I}_{\text{gather}} \leftarrow$ compact $\to$ original (size $N'$; selects representatives)
\STATE \quad $\mathbf{I}_{\text{scatter}} \leftarrow$ original $\to$ compact (size $N$; maps duplicates to unique)
\STATE \quad $\tilde{\mathbf{p}} \leftarrow \mathbf{p}[\mathbf{I}_{\text{gather}}]$ \COMMENT{Compact position IDs for RoPE}

\STATE \textbf{Embedding (Compact Space):}
\STATE $\tilde{\mathbf{x}} \leftarrow \mathbf{x}[\mathbf{I}_{\text{gather}}]$ \COMMENT{Gather token IDs first}
\STATE $\mathbf{h}_0 \leftarrow \text{Embed}(\tilde{\mathbf{x}})$ \COMMENT{Embed unique IDs in compact space}

\STATE \textbf{Position Encoding:}
\STATE $\boldsymbol{\cos} \leftarrow \text{IndexSelect}(\mathbf{C}_{\text{cos}}, \tilde{\mathbf{p}})$
\STATE $\boldsymbol{\sin} \leftarrow \text{IndexSelect}(\mathbf{C}_{\text{sin}}, \tilde{\mathbf{p}})$

\FOR{$\ell = 1$ to $N_{\text{layers}}$}
    \STATE \textbf{Layer $\ell$ Forward:}
    \STATE $\mathbf{h}_{\ell-1}' \leftarrow \text{LayerNorm}(\mathbf{h}_{\ell-1})$
    
    \STATE \textbf{Attention (Causal):}
    \STATE $\mathbf{q}, \mathbf{k}, \mathbf{v} \leftarrow \text{QKVProj}(\mathbf{h}_{\ell-1}')$
    \STATE $\mathbf{q}, \mathbf{k} \leftarrow \text{RoPE}(\mathbf{q}, \mathbf{k}, \boldsymbol{\cos}, \boldsymbol{\sin})$
    
    \STATE \textbf{Expand to Original Layout, and apply flash-attention:}
    \STATE $\mathbf{q}, \mathbf{k}, \mathbf{v} \leftarrow \text{Scatter}(\mathbf{q}, \mathbf{I}_{\text{scatter}}), \text{Scatter}(\mathbf{k}, \mathbf{I}_{\text{scatter}}), \text{Scatter}(\mathbf{v}, \mathbf{I}_{\text{scatter}})$
    
    \STATE $\mathbf{a} \leftarrow \text{FlashAttn}(\mathbf{q}, \mathbf{k}, \mathbf{v}, \mathbf{s}, \text{causal}=True)$
    
    \STATE \textbf{Compact Back:}
    \STATE $\mathbf{a} \leftarrow \text{Gather}(\mathbf{a}, \mathbf{I}_{\text{gather}})$
    \STATE $\mathbf{a} \leftarrow \text{OProj}(\mathbf{a})$
    
    \STATE \textbf{Residual and MLP:}
    \STATE $\mathbf{h}_{\ell}^{\text{attn}} \leftarrow \mathbf{a} + \mathbf{h}_{\ell-1}$ \COMMENT{Post-attention residual}
    \STATE $\mathbf{h}_{\ell}' \leftarrow \text{LayerNorm}(\mathbf{h}_{\ell}^{\text{attn}})$
    \STATE $\mathbf{h}_{\ell} \leftarrow \text{MLP}(\mathbf{h}_{\ell}') + \mathbf{h}_{\ell}^{\text{attn}}$ \COMMENT{Post-MLP residual}
\ENDFOR

\STATE \textbf{Output:}
\STATE $\mathbf{h}_{N_{\text{layers}}} \leftarrow \text{LayerNorm}(\mathbf{h}_{N_{\text{layers}}})$
\STATE $\mathbf{y}_c \leftarrow \text{LMHead}(\mathbf{h}_{N_{\text{layers}}})$
\STATE $\mathbf{y} \leftarrow \text{Scatter}(\mathbf{y}_c, \mathbf{I}_{\text{scatter}})$ \COMMENT{Expand for output}
\RETURN $\mathbf{y}$
\end{algorithmic}
\end{algorithm}

\subsection{Performance: CPU-Side Index Construction Runtime}

Table~\ref{tab:radix_mlp_performance} reports CPU time to build the compaction indices ($I_{gather}$, $I_{scatter}$). In our benchmark, construction stays below a few milliseconds for practical batch sizes and sequence lengths, and can be overlapped with GPU work in an asynchronous scheduler. We also include edge-case fast paths (e.g., a single sequence) that bypass trie construction. 

While the construction time depends on the input shape, it remains low in absolute time. We estimated that the actual inference time for the forward pass will be three orders of magnitude higher. 

All measurements are mean values from 50 samples collected using the rust release build settings, taken directly from \texttt{cargo bench} on a standard Intel server CPU. 
Prefix ratio represents the fraction of each sequence that is shared across all sequences in the batch.

Ultra high batch sizes (e.g., 2048 sequences of 512 tokens) lead to higher absolute index construction times (60.75 ms), but the corresponding prefill time also increases proportionally.
In case this becomes a bottleneck, this can be mitigated by parallelizing the trie construction across multiple CPU threads, and merging the resulting tries.

\begin{table}[htbp]
\centering
\caption{Runtime performance of RadixMLP for creating scatter and gather tensors across different batch sizes and sequence lengths.}
\label{tab:radix_mlp_performance}
\begin{tabular}{l c c c c}
\toprule
\textbf{Configuration} & \textbf{Parameter} & \textbf{Mean Time} & \textbf{Std. Dev.} & \textbf{Tokens/sec} \\
\midrule
\multicolumn{5}{l}{\textit{Batch Size Variation (seq\_len = 512 x \# batches, 0.25 prefix ratio)}} \\
Batch size & 4 & $44.34\,\mu\mathrm{s}$ & $0.05\,\mu\mathrm{s}$ & 46.1M \\
Batch size & 8 & $153.15\,\mu\mathrm{s}$ & $0.10\,\mu\mathrm{s}$ & 26.8M \\
Batch size & 16 & $304.94\,\mu\mathrm{s}$ & $1.44\,\mu\mathrm{s}$ & 26.7M \\
Batch size & 32 & $576.46\,\mu\mathrm{s}$ & $3.46\,\mu\mathrm{s}$ & 28.4M \\
Batch size & 64 & $1.16\,\mathrm{ms}$ & $0.03\,\mathrm{ms}$ & 28.4M \\
Batch size & 256 & $4.68\,\mathrm{ms}$ & $0.02\,\mathrm{ms}$ & 28.0M \\
Batch size & 2048 & $60.75\,\mathrm{ms}$ & $0.19\,\mathrm{ms}$ & 17.3M \\
\midrule
\multicolumn{5}{l}{\textit{Sequence Length Variation (batch\_size = 32, 0.25 prefix ratio)}} \\
Sequence length & 128 & $146.02\,\mu\mathrm{s}$ & $0.25\,\mu\mathrm{s}$ & 28.1M \\
Sequence length & 256 & $295.74\,\mu\mathrm{s}$ & $0.52\,\mu\mathrm{s}$ & 27.8M \\
Sequence length & 512 & $605.01\,\mu\mathrm{s}$ & $1.21\,\mu\mathrm{s}$ & 27.2M \\
Sequence length & 1024 & $1.22\,\mathrm{ms}$ & $0.03\,\mathrm{ms}$ & 27.1M \\
Sequence length & 2048 & $2.29\,\mathrm{ms}$ & $0.05\,\mathrm{ms}$ & 28.7M \\
\midrule
\multicolumn{5}{l}{\textit{Prefix Ratio Variation (batch\_size = 32, seq\_len = 512)}} \\
Prefix ratio & 0.0 & $750.06\,\mu\mathrm{s}$ & $2.00\,\mu\mathrm{s}$ & 22.0M \\
Prefix ratio & 0.25 & $655.93\,\mu\mathrm{s}$ & $18.44\,\mu\mathrm{s}$ & 25.2M \\
Prefix ratio & 0.5 & $455.32\,\mu\mathrm{s}$ & $0.95\,\mu\mathrm{s}$ & 36.3M \\
Prefix ratio & 0.75 & $293.80\,\mu\mathrm{s}$ & $0.23\,\mu\mathrm{s}$ & 56.3M \\
Prefix ratio & 1.0 & $129.98\,\mu\mathrm{s}$ & $0.09\,\mu\mathrm{s}$ & 127.4M \\
\midrule
\multicolumn{5}{l}{\textit{Edge Cases}} \\
Single sequence & 512 & $289.07\,\mathrm{ns}$ & $0.67\,\mathrm{ns}$ & 1.8B \\
Identical sequences & 32$\times$512 & $125.64\,\mu\mathrm{s}$ & $0.07\,\mu\mathrm{s}$ & 130.7M \\
No overlap & 32$\times$512 & $750.41\,\mu\mathrm{s}$ & $1.68\,\mu\mathrm{s}$ & 21.9M \\
\bottomrule
\end{tabular}
\end{table}

We open-source Rust and Python packages under the MIT License, with \texttt{pip install radix\_mlp} or \texttt{cargo add radix\_mlp}.

\subsection{Runtime Analysis: Index-Select Kernel Performance}
\label{sec:appendix:scatter_gather_benchmarks}

To validate the efficiency of our custom gather/scatter kernels, we benchmark \texttt{candle-index-select-cu} against Candle 0.9.1's native index-select implementation on NVIDIA H100. Our kernels demonstrate substantial speedups, particularly for larger tensor shapes typical in transformer inference workloads.

\begin{table}[htbp]
\centering
\small
\caption{Index-Select Kernel Performance: candle-index-select-cu vs Candle 0.9.1 native kernels on NVIDIA H100 80GB HBM3.}
\label{tab:index_select_performance}
\begin{tabular}{l c c c c c}
\toprule
\textbf{Input Shape} & \textbf{Out Rows} & \textbf{DType} & \textbf{Candle 0.9.1} & \textbf{Our Kernels} & \textbf{Speedup} \\
\midrule
$[100, 128]$ & 200 & F32 & 16.436 us & 12.645 us & 1.30 \\
$[100, 128]$ & 200 & F16 & 16.676 us & 12.505 us & 1.33 \\
$[16000, 1024]$ & 12000 & F32 & 340.410 us & 46.250 us & 7.36 \\
$[16000, 1024]$ & 12000 & F16 & 106.553 us & 28.435 us & 3.75 \\
$[16000, 1024]$ & 70000 & F32 & 1.980 ms & 204.540 us & 9.68 \\
$[16000, 1024]$ & 70000 & F16 & 1.073 ms & 106.728 us & 10.06 \\
$[100000, 2048]$ & 500000 & F32 & 45.636 ms & 2.828 ms & 16.14 \\
$[100000, 2048]$ & 500000 & F16 & 33.665 ms & 1.469 ms & 22.92 \\
$[10, 100, 128]$ & 200 & F32 & 33.675 us & 16.217 us & 2.08 \\
$[10, 100, 128]$ & 200 & F16 & 33.953 us & 17.115 us & 1.98 \\
$[2000, 64, 256]$ & 10000 & F32 & 3.737 ms & 432.668 us & 8.64 \\
$[2000, 64, 256]$ & 10000 & F16 & 2.880 ms & 211.468 us & 13.62 \\
\bottomrule
\end{tabular}
\end{table}

The results show that our custom kernels achieve 1.3--22.9× speedups over Candle's native implementation, with the largest gains observed for larger tensor shapes typical in batch transformer inference. This performance improvement is crucial for making RadixMLP's gather/scatter operations negligible compared to the saved position-wise compute.

\subsection{Algorithm: Radix Tree Construction and Index Generation}

Algorithm~\ref{alg:radix_tree} shows a simple CPU procedure for building the prefix trie and producing the index tensors used by RadixMLP. Nodes are path-specific and keyed by the next $(\text{token\_id}, \text{position\_id})$ pair (we combine them into a 64-bit hash via $(position \ll 32) \oplus token$).

\paragraph{Index semantics.}
$I_{gather}$ has length $N'$ and maps compact indices to the first corresponding position in the original layout (compact $\to$ original). $I_{scatter}$ has length $N$ and maps each original position to its compact representative (original $\to$ compact).

\paragraph{Toy example.}
Consider two sequences $[1,2,3]$ and $[1,2,4]$ at positions $[0,1,2]$. Consider the ragged layout of tokens $[1,2,3,1,2,4]$ and \texttt{cu\_seqlens = [0,3,6]}. 
The radix tree reuses the shared prefix $(1,2)$ and creates a new node at the final token.

\begin{table}[h]
\centering
\setlength{\tabcolsep}{4pt}
\renewcommand{\arraystretch}{1.05}
\begin{tabular}{c c c l c}
\toprule
Seq & Pos & Tok & Action & Compact \\
\midrule
1 & 0 & 1 & create & 0 \\
1 & 1 & 2 & create & 1 \\
1 & 2 & 3 & create & 2 \\
2 & 0 & 1 & reuse  & 0 \\
2 & 1 & 2 & reuse  & 1 \\
2 & 2 & 4 & create & 3 \\
\bottomrule
\end{tabular}
\end{table}

\noindent
The resulting index mappings are
$I_{\text{gather}}=[0,1,2,5]$  and
$I_{\text{scatter}}=[0,1,2,0,1,3]$,
yielding 4 unique nodes from 6 tokens (66.7\% compression).

\begin{algorithm}[H]
\caption{Radix Tree Creation}
\label{alg:radix_tree}
\begin{algorithmic}[1]
\STATE $nodes \leftarrow [Node(root)]$
\STATE $next\_compact \leftarrow 0$
\STATE $scatter\_indices \leftarrow []$ \COMMENT{Size will be $N$ (total original tokens)}
\STATE $gather\_indices \leftarrow []$ \COMMENT{Size will be $N'$ (unique compact tokens)}
\FOR{$s = 0$ \textbf{to} $num\_sequences - 1$}
    \STATE $start \leftarrow cu\_seq\_lengths[s]$
    \STATE $end \leftarrow cu\_seq\_lengths[s+1]$
    \STATE $parent \leftarrow 0$
    \FOR{$i = start$ \textbf{to} $end - 1$}
        \STATE $token \leftarrow input\_ids[i]$
        \STATE $position \leftarrow position\_ids[i]$
        \STATE $key \leftarrow (position \ll 32) \oplus token$
        \STATE $found \leftarrow \textsc{False}$
        \STATE $compact\_idx \leftarrow -1$
        \STATE $children \leftarrow nodes[parent].children$
        \FOR{$j = 0$ \textbf{to} $children.length - 1$}
            \IF{$children[j].key == key$}
                \STATE $parent \leftarrow children[j].node\_idx$
                \STATE $compact\_idx \leftarrow nodes[parent].compact\_index$
                \STATE $found \leftarrow \textsc{True}$
                \STATE \textbf{break}
            \ENDIF
        \ENDFOR
        \IF{$\neg found$}
            \STATE $idx \leftarrow nodes.length$
            \STATE $nodes[idx] \leftarrow Node(key, s, i-start)$
            \STATE $compact\_idx \leftarrow next\_compact$
            \STATE $nodes[idx].compact\_index \leftarrow compact\_idx$
            \STATE $next\_compact \leftarrow next\_compact + 1$
            \STATE $nodes[parent].children.append((key, idx))$
            \STATE $parent \leftarrow idx$
            \STATE $gather\_indices.append(i)$ \COMMENT{Map compact $\to$ first original}
        \ENDIF
        \STATE $scatter\_indices.append(compact\_idx)$ \COMMENT{Map original $\to$ compact}
    \ENDFOR
\ENDFOR
\RETURN $RadixTree(nodes, scatter\_indices, gather\_indices)$
\end{algorithmic}
\end{algorithm}

\subsection{Optimization: Performance Padding for GPU Utilization}
\label{sec:performance_padding}

To improve GPU utilization and enable CUDA Graph capture, we provide an optional implementation that pads the compact token count $N'$ to a fixed bucket size by repeating the first entry of $I_{gather}$. 
The padded outputs are discarded after computation, so results are unchanged; the trade-off is a small amount of extra position-wise compute that can reduce graph recompilations and improve utilization at aligned tensor sizes.
As example, many models are padding their embedding layer vocab size to a multiple of 64.
We disable performance padding and CUDA Graph capture for all experiments. By default, most frameworks do not enable CUDA Graphs for the prefill phase.
As described in Algorithm~\ref{alg:radix_mlp_causal_lm}, this padding applies to all position-wise operations, including the $Q$, $K$, $V$, and $O$ projections, Normalization Layers, and RoPE computations.

\section{Training Validation: Numerical Equivalence Analysis}
\label{sec:appendix:training}

This appendix evaluates numerical equivalence of RadixMLP in training mode (forward and backward) and separates effects due to (i) RadixMLP compaction and (ii) the attention backend.

\begin{table*}[th]
\centering
\scriptsize
\setlength{\tabcolsep}{4.5pt}
\renewcommand{\arraystretch}{1.15}
\caption{\textbf{Forward/backward numerical differences vs.\ SDPA + no\_radix (reference).}
We report $(\max|\Delta|,\;\mathrm{mean}|\Delta|)$ for logits (forward) and parameter gradients (backward). As patterns, we test all sequences identical, shared prefixes, no sharing, mixed lengths, and complex/arbitrary trie branching.
The three panels isolate: (top) RadixMLP effect under SDPA, (middle) backend swap under no\_radix, (bottom) backend swap under radix. All ablations compare against no\_radix SDPA baseline. }
\label{tab:radix_mlp_attention_ablation_threepanel}

% --- Panel 1: SDPA-only (RadixMLP effect)
\begin{subtable}{\textwidth}
\label{tab:train_radix_mlp_effect}
\centering
\caption{\textbf{SDPA backend: RadixMLP toggle (isolates RadixMLP).}}
\begin{tabular}{@{}l c cc@{}}
\toprule
\textbf{Test case} & \textbf{\#tokens} &
\textbf{SDPA + radix logits} $(\max,\mathrm{mean})$ &
\textbf{SDPA + radix grads} $(\max,\mathrm{mean})$ \\
\midrule
single\_sequence*     & 5  & $(0.0,\;0.0)$ & $(0.0,\;0.0)$ \\
identical\_sequences & 10 & $(0.0,\;0.0)$ & $(3.81{\times}10^{-6},\;3.0{\times}10^{-8})$ \\
shared\_prefix       & 10 & $(0.0,\;0.0)$ & $(1.144{\times}10^{-5},\;1.0{\times}10^{-7})$ \\
no\_sharing*          & 6  & $(0.0,\;0.0)$ & $(0.0,\;0.0)$ \\
mixed\_lengths       & 10 & $(0.0,\;0.0)$ & $(1.144{\times}10^{-5},\;1.5{\times}10^{-7})$ \\
complex\_sharing     & 20 & $(4.8{\times}10^{-7},\;9.0{\times}10^{-8})$ & $(1.907{\times}10^{-5},\;3.9{\times}10^{-7})$ \\
\bottomrule
\end{tabular}
\end{subtable}

\vspace{0.75em}

% --- Panel 2: Backend swap with no_radix
\begin{subtable}{\textwidth}
\label{tab:train_backend_swap_effect}
\centering
\caption{\textbf{Backend swap with no\_radix: FA2 vs.\ SDPA (isolates backend effect).}}
\begin{tabular}{@{}l c cc@{}}
\toprule
\textbf{Test case} & \textbf{\#tokens} &
\textbf{FA2 + no\_radix logits} $(\max,\mathrm{mean})$ &
\textbf{FA2 + no\_radix grads} $(\max,\mathrm{mean})$ \\
\midrule
single\_sequence     & 5  & $(2.1654{\times}10^{-4},\;4.271{\times}10^{-5})$ & $(1.512146{\times}10^{-2},\;1.4903{\times}10^{-4})$ \\
identical\_sequences & 10 & $(2.1654{\times}10^{-4},\;4.271{\times}10^{-5})$ & $(3.024101{\times}10^{-2},\;2.9815{\times}10^{-4})$ \\
shared\_prefix       & 10 & $(2.1654{\times}10^{-4},\;4.216{\times}10^{-5})$ & $(1.653290{\times}10^{-2},\;2.4185{\times}10^{-4})$ \\
no\_sharing          & 6  & $(2.2042{\times}10^{-4},\;4.518{\times}10^{-5})$ & $(1.379967{\times}10^{-2},\;1.5119{\times}10^{-4})$ \\
mixed\_lengths       & 10 & $(2.1654{\times}10^{-4},\;4.484{\times}10^{-5})$ & $(2.959824{\times}10^{-2},\;2.7050{\times}10^{-4})$ \\
complex\_sharing     & 20 & $(2.1660{\times}10^{-4},\;3.974{\times}10^{-5})$ & $(3.052521{\times}10^{-2},\;3.6787{\times}10^{-4})$ \\
\bottomrule
\end{tabular}
\end{subtable}

\vspace{0.75em}

% --- Panel 3: Backend swap with radix
\begin{subtable}{\textwidth}
\label{tab:train_backend_swap_radix_mlp_effect}
\centering
\caption{\textbf{Backend swap with radix: FA2 vs.\ SDPA (backend + RadixMLP effects).}}
\begin{tabular}{@{}l c cc@{}}
\toprule
\textbf{Test case} & \textbf{\#tokens} &
\textbf{FA2 + radix logits} $(\max,\mathrm{mean})$ &
\textbf{FA2 + radix grads} $(\max,\mathrm{mean})$ \\
\midrule
single\_sequence     & 5  & $(2.1654{\times}10^{-4},\;4.271{\times}10^{-5})$ & $(1.512146{\times}10^{-2},\;1.4903{\times}10^{-4})$ \\
identical\_sequences & 10 & $(2.1654{\times}10^{-4},\;4.271{\times}10^{-5})$ & $(3.023911{\times}10^{-2},\;2.9816{\times}10^{-4})$ \\
shared\_prefix       & 10 & $(2.1654{\times}10^{-4},\;4.216{\times}10^{-5})$ & $(2.268410{\times}10^{-2},\;2.7209{\times}10^{-4})$ \\
no\_sharing          & 6  & $(2.2042{\times}10^{-4},\;4.518{\times}10^{-5})$ & $(1.379967{\times}10^{-2},\;1.5119{\times}10^{-4})$ \\
mixed\_lengths       & 10 & $(2.1654{\times}10^{-4},\;4.484{\times}10^{-5})$ & $(2.085495{\times}10^{-2},\;2.7642{\times}10^{-4})$ \\
complex\_sharing     & 20 & $(2.1666{\times}10^{-4},\;3.975{\times}10^{-5})$ & $(3.616333{\times}10^{-2},\;3.9834{\times}10^{-4})$ \\
\bottomrule
\end{tabular}
\end{subtable}

\end{table*}

\paragraph{Experimental Setup.}
We test a 2-layer Qwen3-style model (hidden size 256, intermediate size 512) on synthetic variable-length batches with different prefix-sharing patterns (following the setup described in \cref{sec:experiments}). We enable CUDA deterministic settings and disable TF32. We compare four configurations formed by the Cartesian product of RadixMLP toggle \{off,on\} and attention backend \{FA2, SDPA \}.

For each test case, we report maximum and mean absolute differences in logits and gradients relative to a single reference configuration. We design a set of test cases, where all tokens are duplicated across all batch slots (identical\_sequences), have a shared prefix but then branch off (shared\_prefix), have zero common tokens (no\_sharing), have tokens of mixed lengths, or represent a complex sharing scenario with varying numbers of mutually shared tokens. 
In case of trivial cases* (single\_sequence, no\_sharing), we skip the RadixMLP operations altogether in the computational graph. RadixMLP has zero effect in these cases, as it's unused.
Operations that are stochastic, such as dropout are not enabled. Adding dropout would not be conceptually possible with this setup, as dropout in compact tokens would affect multiple original tokens, breaking numerical equivalence. Such operations however are no longer typically used in modern transformer training (e.g. Qwen3 / Llama3 do not use dropout).

\paragraph{Reference.}
Unless stated otherwise, the reference is \textbf{SDPA + no\_radix}. This isolates (a) the effect of RadixMLP under the same attention backend (SDPA, ~\cref{tab:train_radix_mlp_effect}a), and (b) the effect of swapping attention backends (FA2 vs.\ SDPA,  ~\cref{tab:train_backend_swap_effect}b), and (c) with RadixMLP and FA2 backend (~\cref{tab:train_backend_swap_radix_mlp_effect}c).

\paragraph{Takeaway.}
Across all tested shapes, the differences induced by toggling RadixMLP under a fixed attention backend (SDPA) remain at or below $\sim 2\times 10^{-5}$ in maximum gradient deviation, while switching the attention backend (FA2 vs.\ SDPA) produces substantially larger deviations (up to $\sim 3\times 10^{-2}$ in maximum gradient deviation), independent of whether RadixMLP is enabled.

\section{Additional details for end-to-end benchmark results}
\subsection{Ablations of RadixMLP enable/disable across max-batch-tokens}
\label{sec:appendix:end-to-end-results}

When running the latency results, we used performance-client (\url{https://pypi.org/project/baseten-performance-client/}), an HTTP package that allows fine-grained control over batching parameters. We adjust the settings to simulate the following scenario. 
We divide the MS~MARCO v1.1 validation set of 82{,}360 passages, apply the Qwen3 reranker chat template, and split them into micro-batches of 64 sentences per request. To avoid creating excessive backpressure on the server side, we limit concurrency to 32 in-flight requests (2048 sentences). 

We run all configurations on separate GPUs at the same time. The benchmarking setup is fully reproducible in the repo $^\dagger$ under \texttt{./benchmarks}.

\begin{figure}[h] 
\centering
\begin{subfigure}{\linewidth}
  \centering
  \includegraphics[width=\linewidth,height=.27\textheight,keepaspectratio]{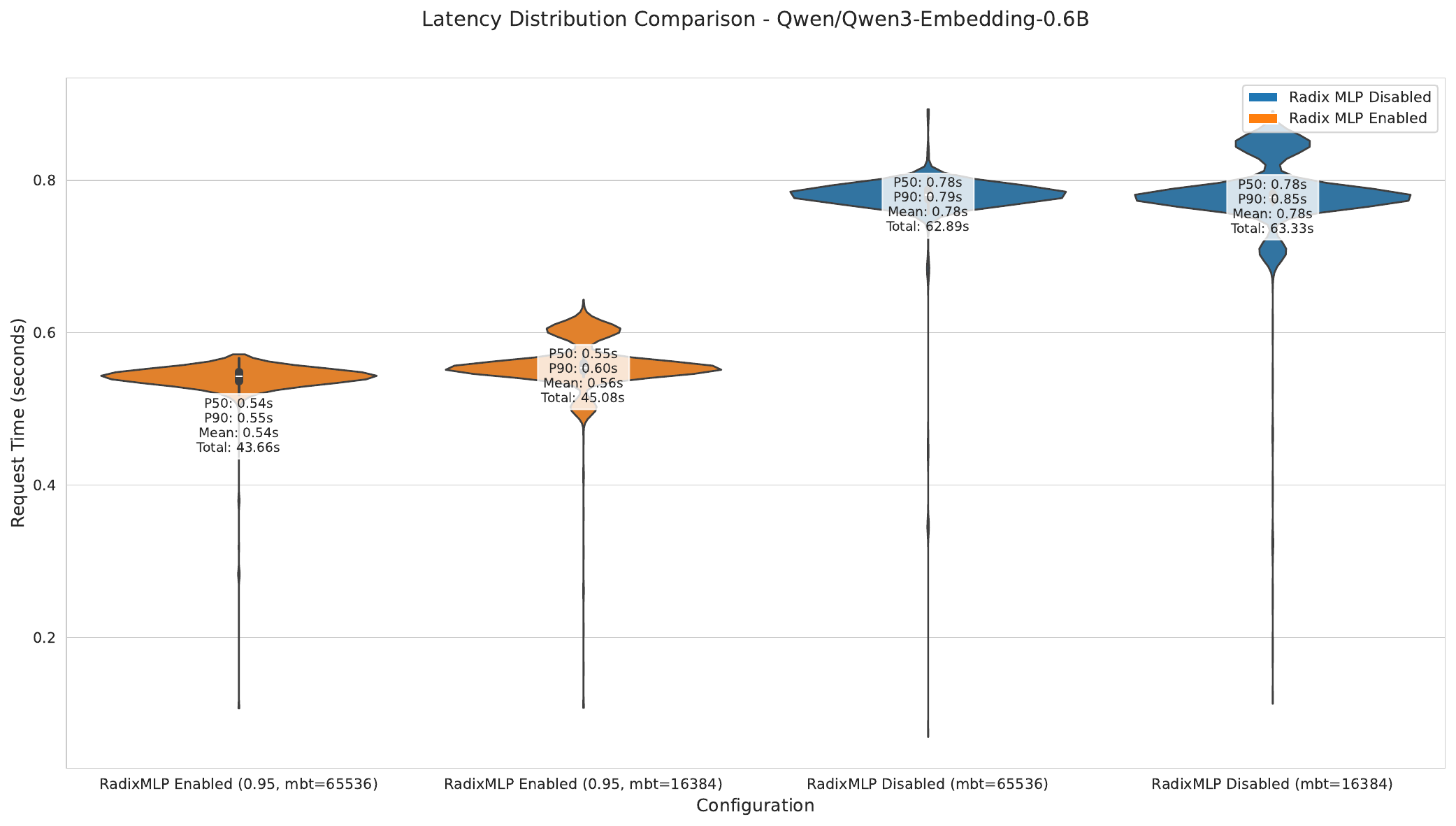}
  \caption{MS~MARCO Benchmark 0.6B}
  \label{fig:radixmlp_06b}
\end{subfigure}

\vspace{1mm}

\begin{subfigure}{\linewidth}
  \centering
  \includegraphics[width=\linewidth,height=.27\textheight,keepaspectratio]{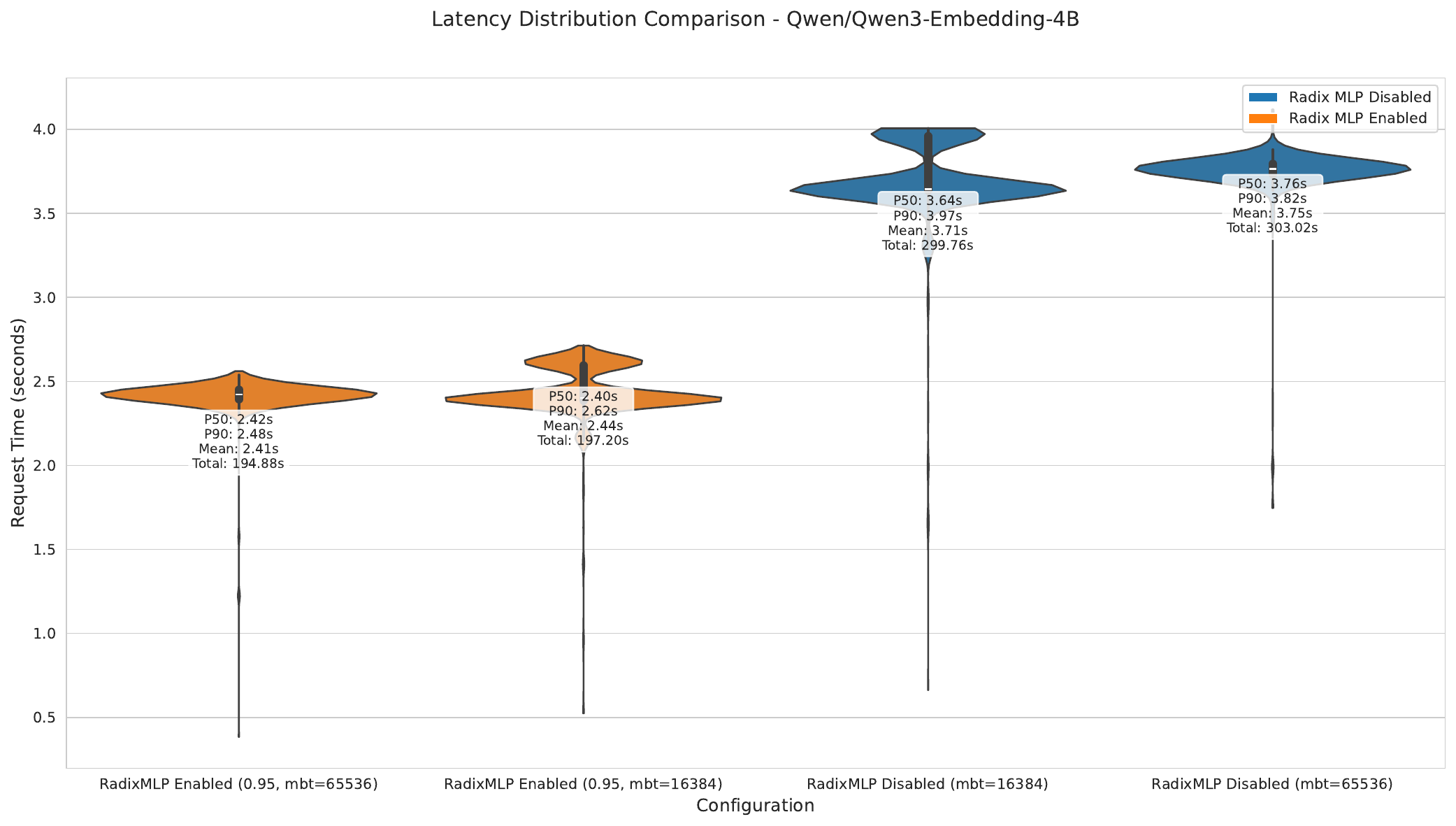}
  \caption{MS~MARCO Benchmark 4B}
  \label{fig:radixmlp_4b}
\end{subfigure}

\vspace{1mm}

\begin{subfigure}{\linewidth}
  \centering
  \includegraphics[width=\linewidth,height=.27\textheight,keepaspectratio]{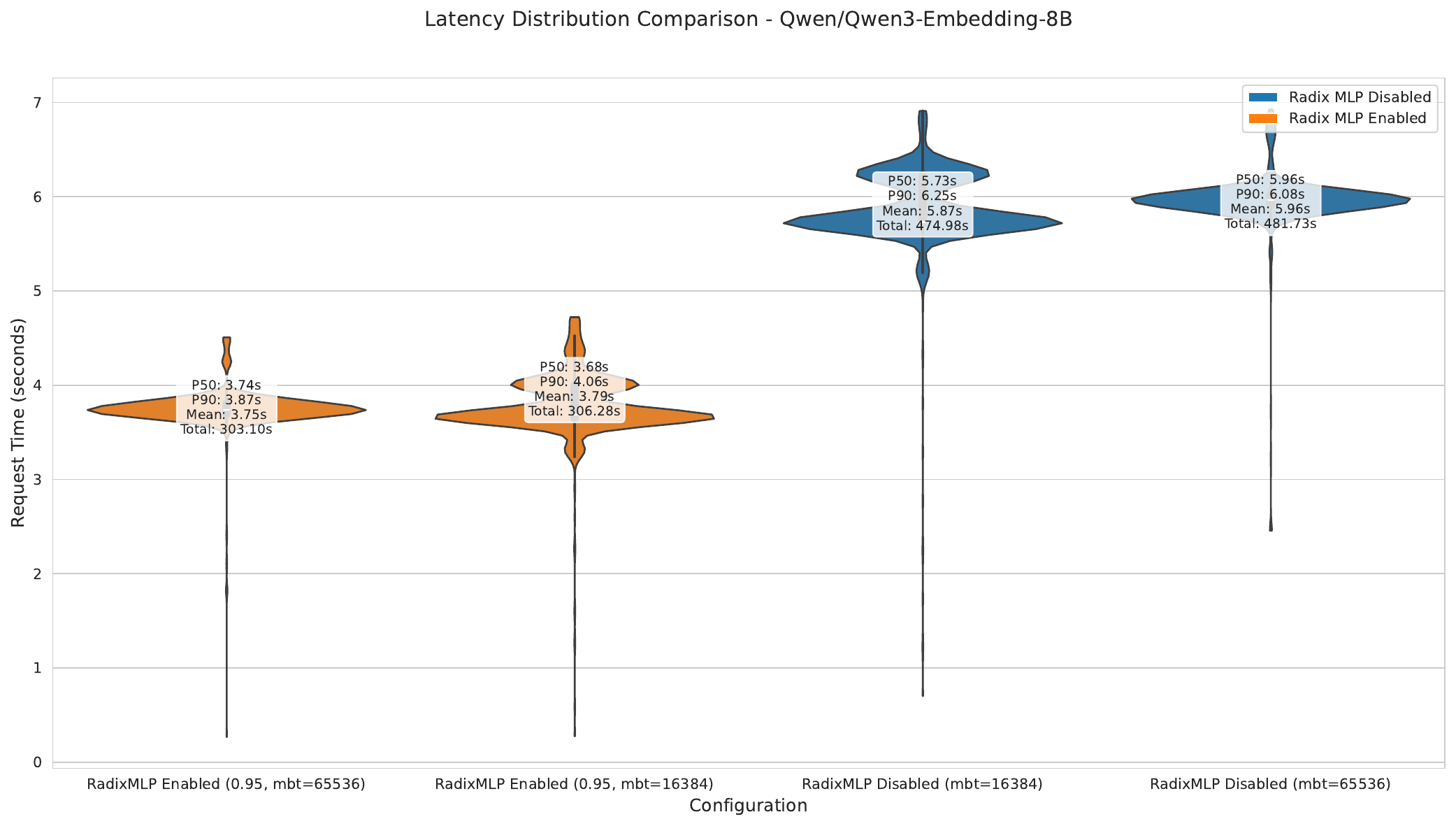}
  \caption{MS~MARCO Benchmark 8B}
  \label{fig:radixmlp_8b}
\end{subfigure}

\caption{\textbf{MS~MARCO latency distribution}: When running separate 1287 requests against TEI, response latency can vary based on the total number of tokens in a request. To show the Pareto improvement of RadixMLP, the three plots show two different settings (RadixMLP enabled/disabled and max batch tokens 16384 or 65536) and their impact on the perceived throughput and latency of a deployment. }
\label{fig:radixmlp_all}
\end{figure}

\subsection{Benchmark: End-to-end Performance Comparison against vLLM}

\label{sec:appendix:end-to-end-results-vLLM}
\begin{figure}[h] 
\centering
\begin{subfigure}{\linewidth}
  \centering
  \includegraphics[width=\linewidth,height=.28\textheight,keepaspectratio]{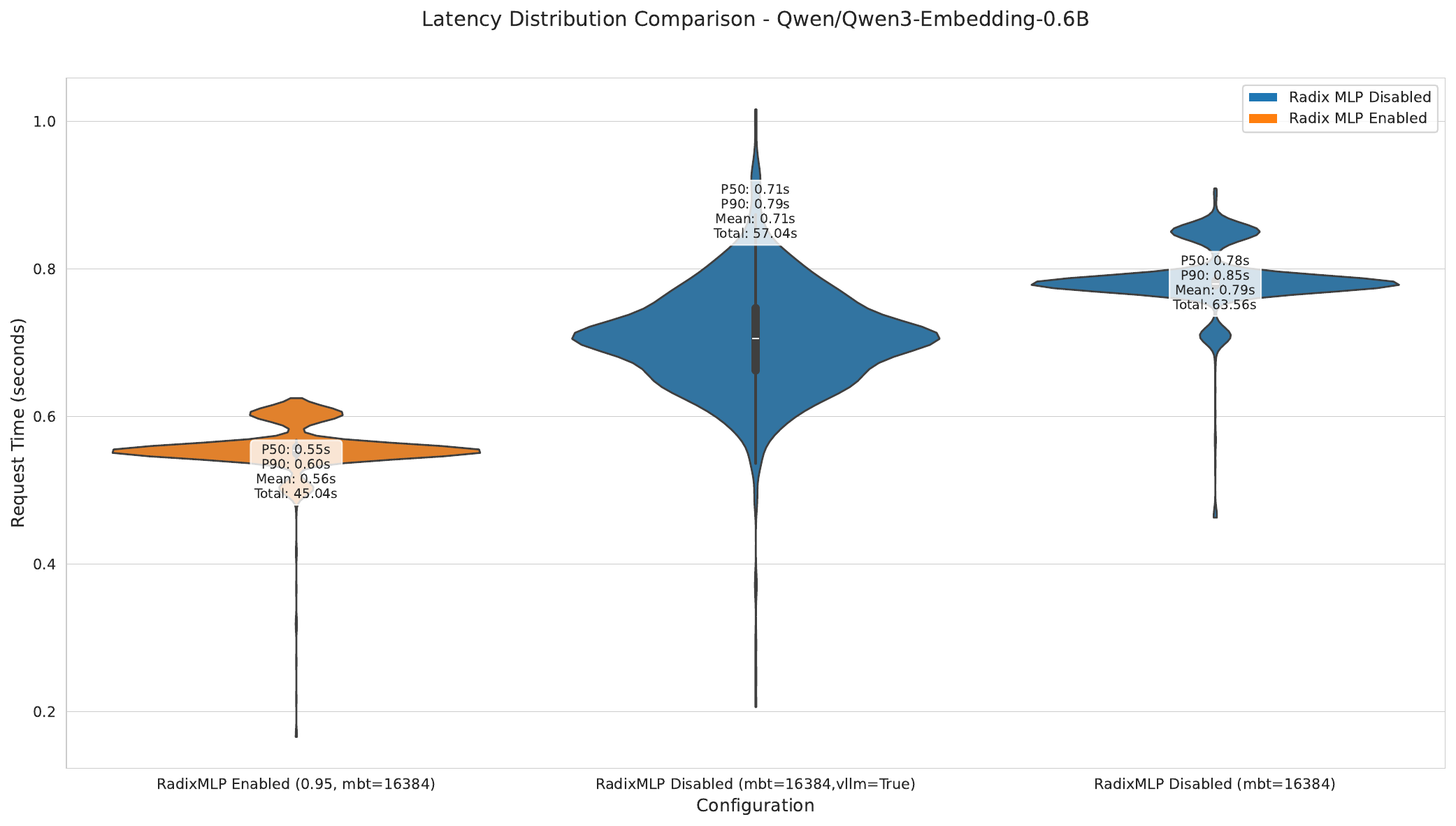}
  \caption{MS~MARCO Benchmark vLLM 0.6B}
  \label{fig:radixmlp_vllm_06b}
\end{subfigure}

\vspace{1mm}

\begin{subfigure}{\linewidth}
  \centering
  \includegraphics[width=\linewidth,height=.28\textheight,keepaspectratio]{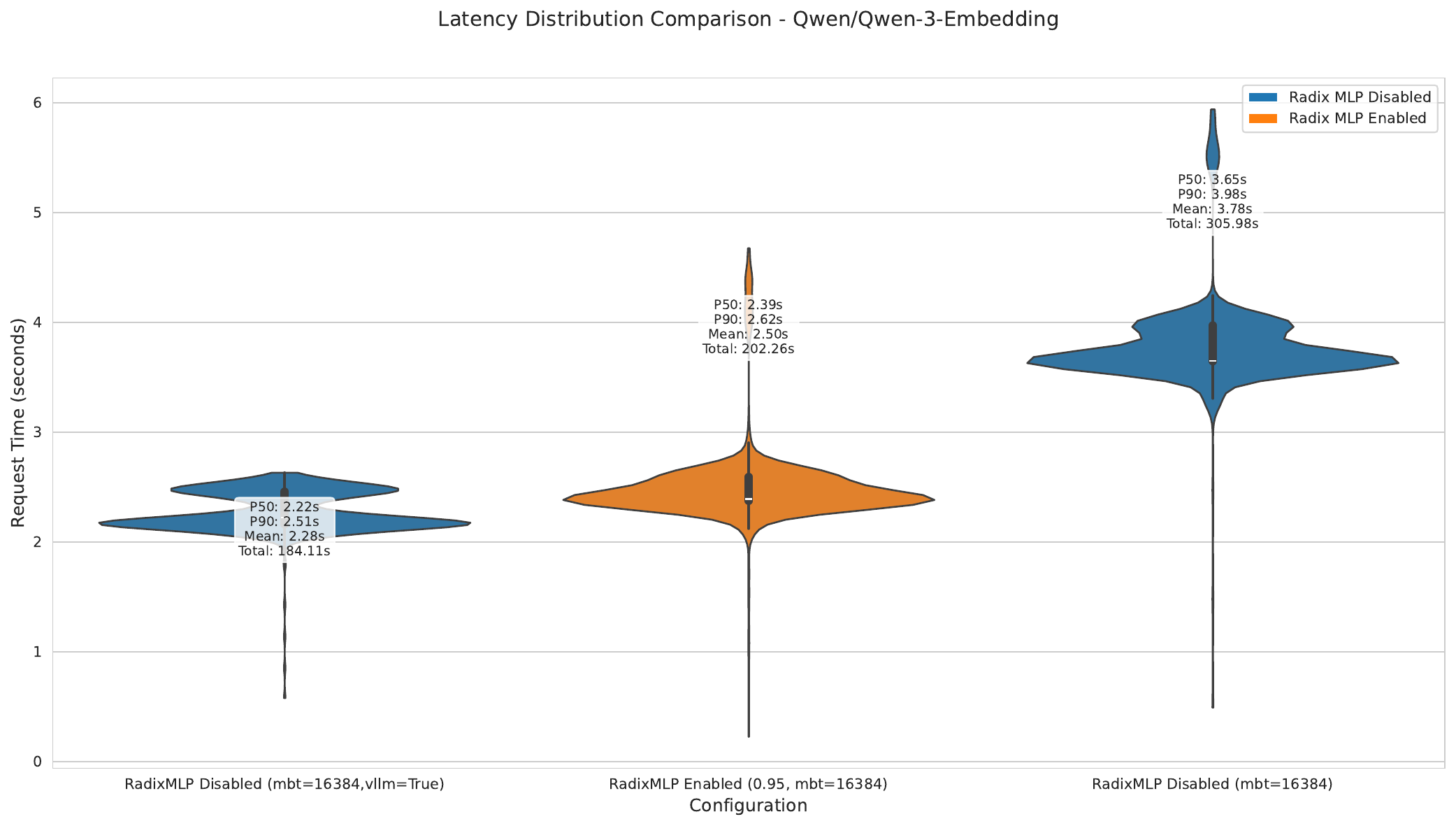}
  \caption{MS~MARCO Benchmark vLLM 4B}
  \label{fig:radixmlp_vllm_4b}
\end{subfigure}

\vspace{1mm}

\begin{subfigure}{\linewidth}
  \centering
  \includegraphics[width=\linewidth,height=.28\textheight,keepaspectratio]{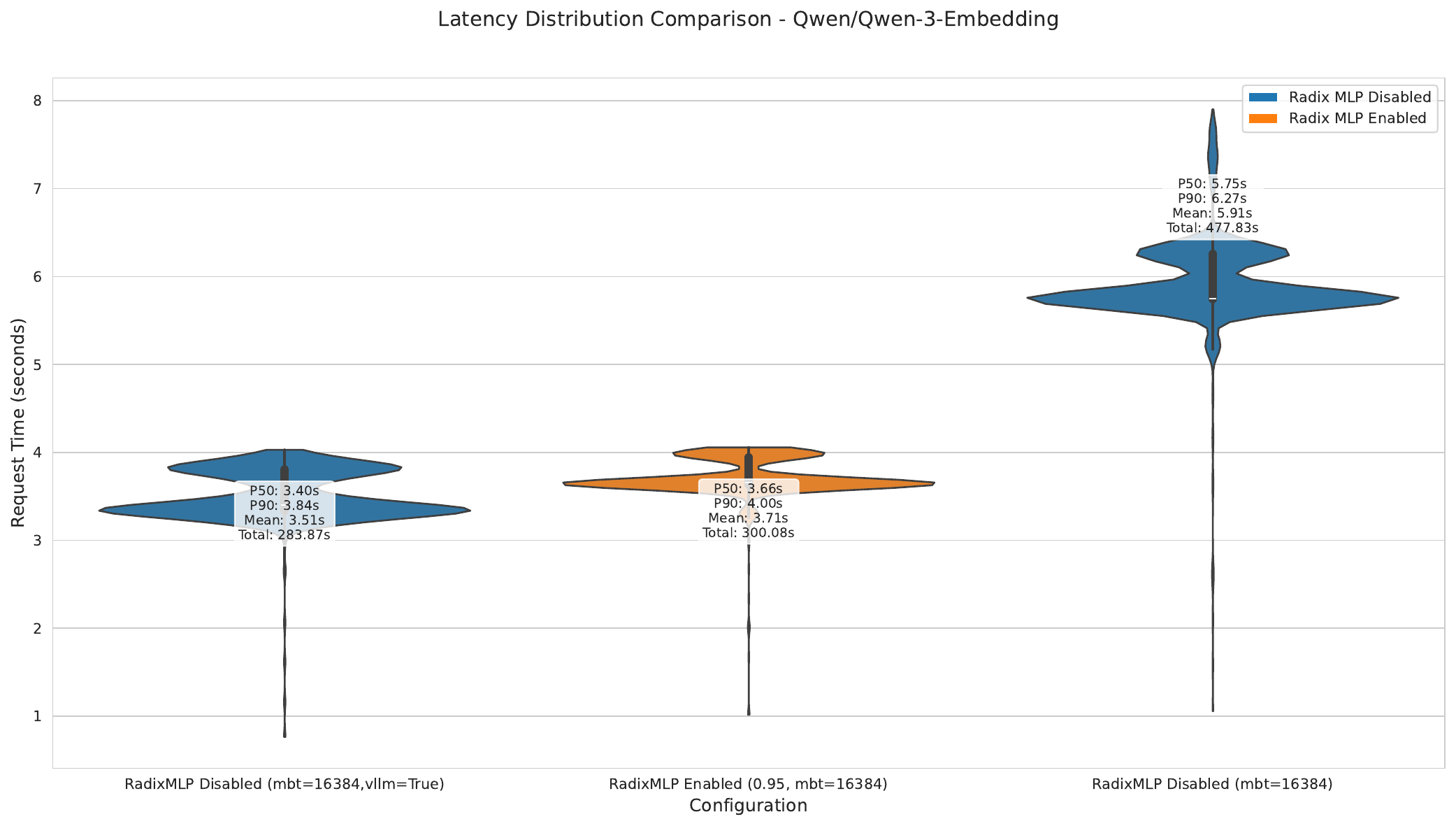}
  \caption{MS~MARCO Benchmark vLLM 8B}
  \label{fig:radixmlp_vllm_8b}
\end{subfigure}

\caption{\textbf{Comparison of TEI against vLLM against the dataset b (validation, regular order)}: With a RadixMLP compression ratio of around 0.61 and a reported KV Cache reuse ratio of 0.31, vLLM is noticeably faster than vanilla TEI across all configurations. Using RadixMLP shows on-par or improved performance over vLLM. }
\label{fig:radixmlp_all_vLLM}
\end{figure}

In \cref{fig:radixmlp_all_vLLM} we perform the same experiments as \cref{fig:radixmlp_all}, by running the validation dataset of MSMarco. 

In its vanilla configuration, we do not expect TEI to be faster. In particular, vLLM has practical advantages such as the use of flash-attention-3 on Hopper, more optimized matmul, layernorm kernels. In comparison, TEI has fewer contributors and still uses flash-attention-2 kernels and defaults to cuBLAS-lt kernels. 
While vLLM's GIL utilization is improved in various pull requests, TEI's web server and tokenization are written in Rust, making it faster for models with high CPU overhead. 

\begin{figure}[h] 
\centering
\begin{subfigure}{\linewidth}
  \centering
  \includegraphics[width=\linewidth,height=.59\textheight,keepaspectratio]{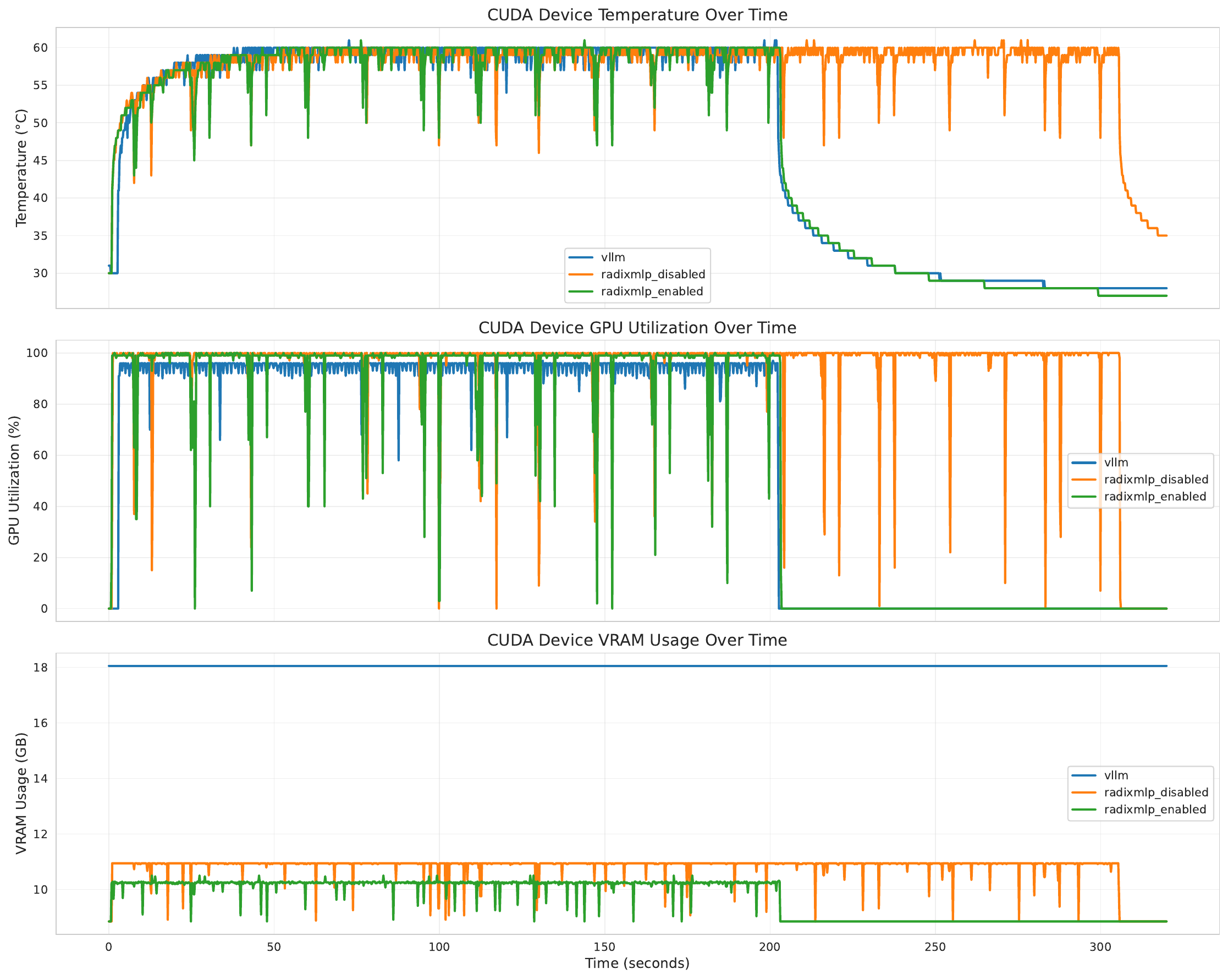}
  \caption{MS~MARCO Benchmark 4B GPU and VRAM Utilization and temperature over time.}
  \label{fig:radixmlp_util1}
\end{subfigure}

\vspace{1mm}

\begin{subfigure}{\linewidth}
  \centering
  \includegraphics[width=\linewidth,height=.32\textheight,keepaspectratio]{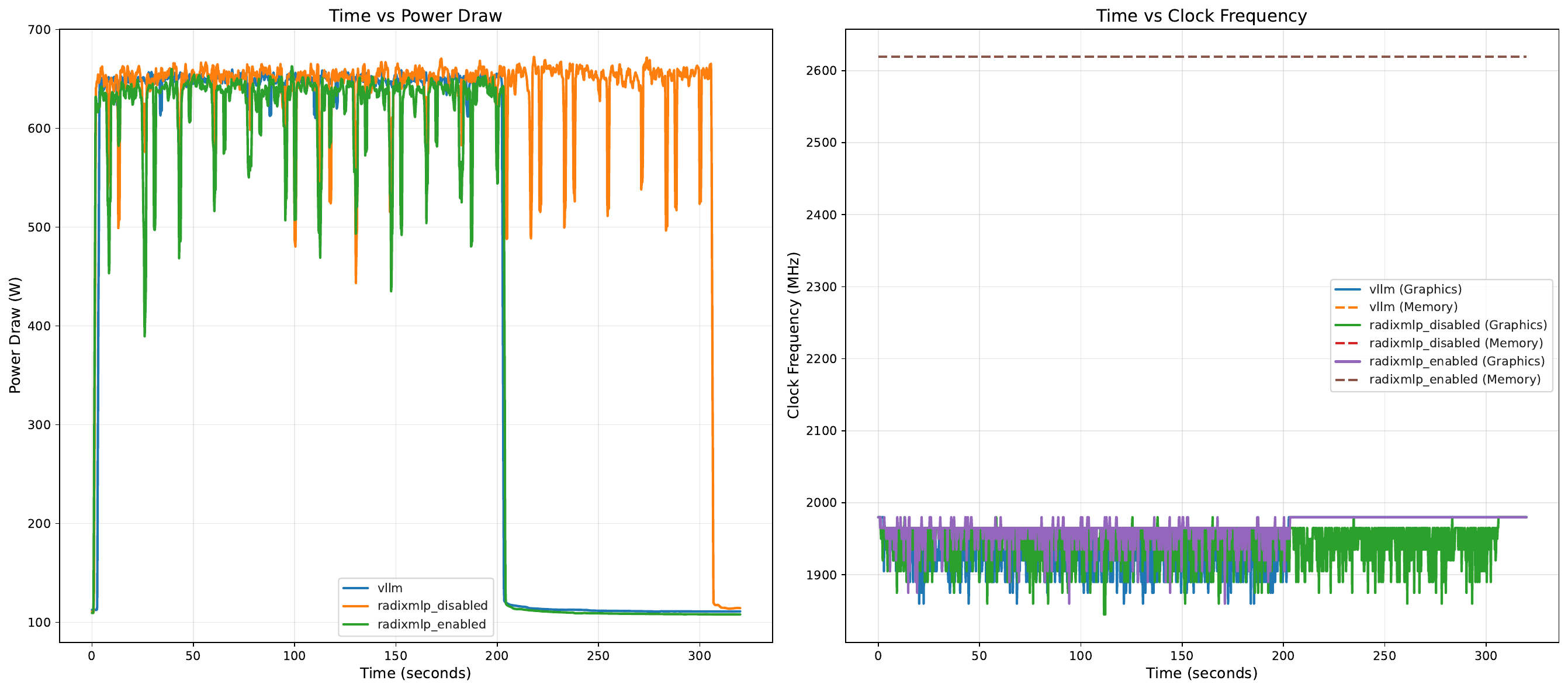}
  \caption{MS~MARCO Benchmark 4B Power draw and clock frequency over time.}
\end{subfigure}

\caption{\textbf{MS~MARCO utilization of TEI and vLLM during the benchmarks  of running 4B models}.}
\label{fig:radixmlp_utilization_vllm}
\end{figure}

Comparing the utilization of GPU resources, \cref{fig:radixmlp_utilization_vllm} shows that all GPUs exhibit a similar power draw and temperature during one of the experiments (recorded over \cref{fig:radixmlp_vllm_4b}). A comparison of VRAM usage is interesting. vLLM was started with its lowest possible memory fraction, leading to a permanent occupation of 18GB of VRAM. For the 4B model, TEI uses a static utilization of 8.5GB VRAM. Without RadixMLP, additional 2.5GB are allocated on average for activation memory. With RadixMLP, and a mean compression ratio of 0.61 for this dataset, the activation memory is reduced to 1.5GB of additional memory.

% comparison with train dataset and reversed

\begin{figure}[h] 
\centering
\begin{subfigure}{\linewidth}
  \centering
  \includegraphics[width=\linewidth,height=.28\textheight,keepaspectratio]{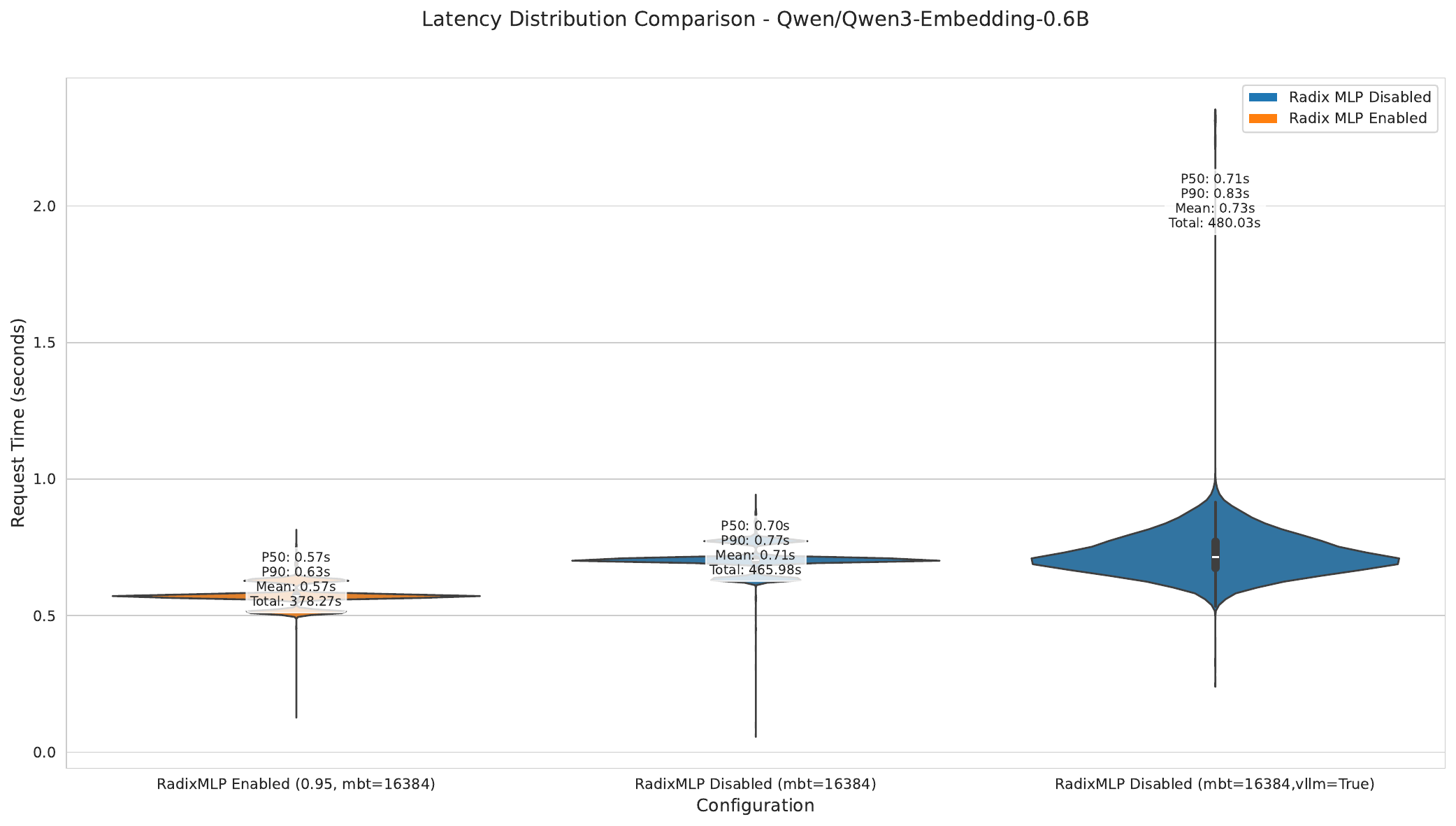}
  \caption{MS~MARCO Benchmark vLLM 0.6B}
  \label{fig:radixmlp_rev_train_06b}
\end{subfigure}

\vspace{1mm}

\begin{subfigure}{\linewidth}
  \centering
  \includegraphics[width=\linewidth,height=.28\textheight,keepaspectratio]{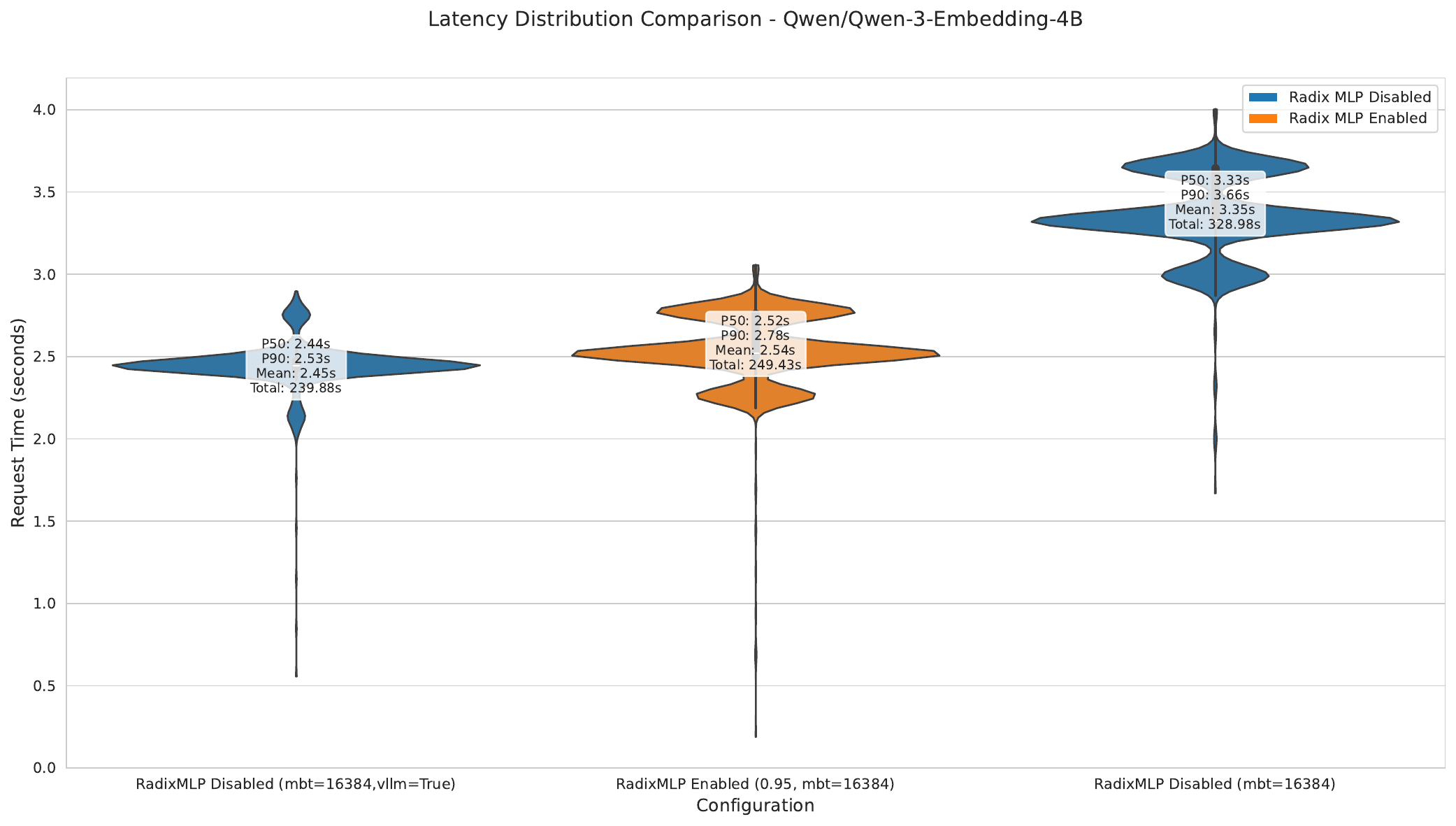}
  \caption{MS~MARCO Benchmark vLLM 4B}
  \label{fig:radixmlp_rev_train_4b}
\end{subfigure}

\vspace{1mm}

\begin{subfigure}{\linewidth}
  \centering
  \includegraphics[width=\linewidth,height=.28\textheight,keepaspectratio]{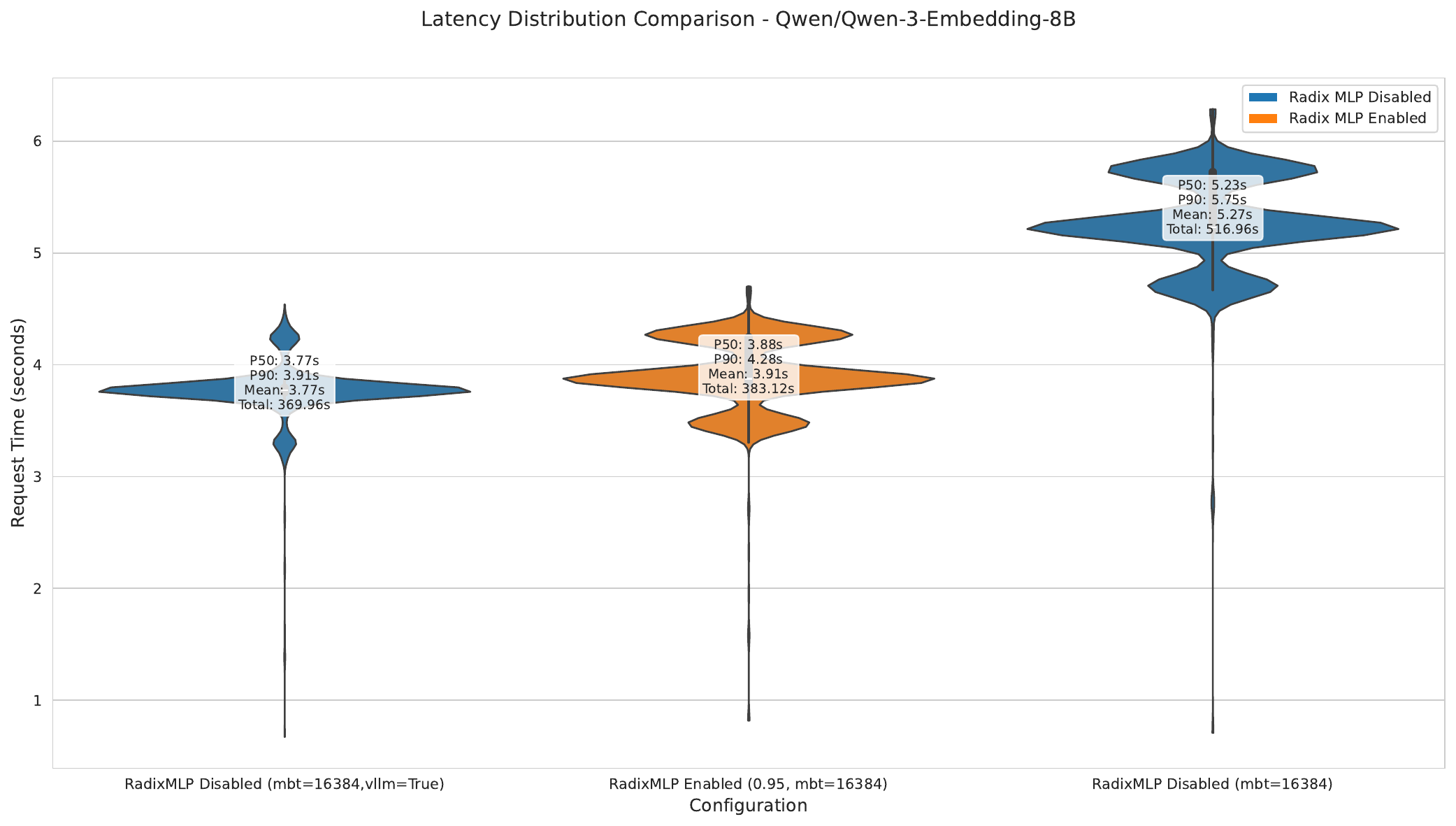}
  \caption{MS~MARCO Benchmark vLLM 8B}
  \label{fig:radixmlp_rev_train_8b}
\end{subfigure}

\caption{\textbf{Comparison of TEI against vLLM against the dataset c (training, flipped order)}: With a lower RadixMLP compression ratio of around 0.71 and a reported KV cache reuse ratio of 0.20, vLLM is again faster than vanilla TEI across all configurations. Using RadixMLP shows on-par or improved performance over vLLM. }
  \label{fig:radixmlp_all_vLLM_rev_train}
\end{figure}

\textbf{Future work} could explore integrating RadixMLP into vLLM, by further deduplicating activations that are left after vLLM's KV cache reuse. This would make vLLM more robust to cache misses and give levers to improve performance by intra-batch deduplication complementary to KV cache reuse. 
However, this integration is non-trivial. At the minimum, it would require adapting vLLM's internal data structures to support the ragged layout for the position-wise operations and support across various features, such as CUDA-graph capture, multi-GPU settings, and require a deep scheduler integration.
For this reason, we leave this exploration to future work.

\end{document}